\def\ACCEPTREVISIONS{1}
\documentclass{article}

\PassOptionsToPackage{sort&compress}{natbib}

\usepackage{iclr2027_conference,times}

\IfFileExists{pcrr8t.tfm}{}{}

\usepackage[utf8]{inputenc}
\usepackage[T1]{fontenc}
\usepackage{graphicx}
\usepackage{booktabs}
\usepackage{tabularx}
\usepackage{longtable}
\newcolumntype{Y}{>{\raggedright\arraybackslash}X}
\newcolumntype{L}[1]{>{\raggedright\arraybackslash}p{#1}}
\usepackage{amsmath,amssymb}
\usepackage{xcolor}
\usepackage[normalem]{ulem}
\usepackage{url}
\usepackage{natbib}
\usepackage{hyperref}
\hypersetup{hidelinks}
\IfFileExists{fontawesome5.sty}{\usepackage{fontawesome5}}{%
  \providecommand{\faEnvelope}{Email:}\providecommand{\faGithub}{Code:}}

\newif\ifanon
\ifdefined\ANONSUBMISSION \anontrue \else \anonfalse \fi

\newcommand{\benchname}{CraftBench-UE}
\ifanon
  \newcommand{\productname}{MCP2}
  
  \newcommand{\artifacturl}{the public project repository}
\else
  \newcommand{\productname}{Aura}
  \newcommand{\artifacturl}{\url{https://github.com/ramenvr/craftbench-ue}}
\fi

\newif\iftrackchanges
\ifdefined\ACCEPTREVISIONS\trackchangesfalse\else\trackchangestrue\fi
\definecolor{revisiontwo}{HTML}{005A9C}
\makeatletter
\DeclareRobustCommand{\revadd}{\@ifnextchar2{\cb@revaddtwo}{\cb@revaddone}}
\DeclareRobustCommand{\revdel}{\@ifnextchar2{\cb@revdeltwo}{\cb@revdelone}}
\long\def\cb@revaddtwo2#1{\cb@addtwo{#1}}
\long\def\cb@revdeltwo2#1{\cb@deltwo{#1}}
\iftrackchanges
  \newcommand{\cb@revdelone}[1]{\textcolor{gray}{\sout{#1}}}
  \newcommand{\cb@revaddone}[1]{\textcolor{red}{#1}}
  \newcommand{\cb@addtwo}[1]{\textcolor{revisiontwo}{#1}}
  \newcommand{\cb@deltwo}[1]{\textcolor{revisiontwo}{\sout{#1}}}
\else
  \newcommand{\cb@revdelone}[1]{}
  \newcommand{\cb@revaddone}[1]{#1}
  \newcommand{\cb@addtwo}[1]{#1}
  \newcommand{\cb@deltwo}[1]{}
\fi
\makeatother

\newcommand{\numtasks}{70}
\newcommand{\numcpp}{33}
\newcommand{\numbp}{25}
\newcommand{\numpython}{12}
\newcommand{\numextensiontasks}{47}

\newcommand{\nummatchedpairs}{10}
\newcommand{\numthirdperson}{50}
\newcommand{\numtemplateproj}{20}
\newcommand{\numpanelmodels}{seven}

\newcommand{\scoredminutes}{40}

\newif\ifresultsfrozen
\resultsfrozenfalse

\newcommand{\armmcp}{MCP1}
\newcommand{\armmcpfullname}{Unreal MCP}
\newcommand{\armaura}{MCP2}
\newcommand{\armaurafullname}{\ifanon an editor tool developed by the authors' organization\else Aura MCP\fi}

\newcommand{\armcontrol}{File/shell baseline}
\newcommand{\ue}{Unreal Engine~5.8}
\newcommand{\pie}{PIE}

\IfFileExists{tables/macros_generated.tex}{% GENERATED by tools/paper/make_tables.py -- do not hand-edit.
}{}
\newcommand{\auditvalue}[1]{\csname cb#1\endcsname}
\expandafter\def\csname cbBpConfigN\endcsname{350}
\expandafter\def\csname cbBpConfigprovidersserved\endcsname{0}
\expandafter\def\csname cbBpConfigreasoningpolicy\endcsname{272}
\expandafter\def\csname cbBpConfigsystempromptsha\endcsname{0}
\expandafter\def\csname cbBpConfigtoolsoffered\endcsname{0}
\expandafter\def\csname cbBpDeclaredL1\endcsname{25}
\expandafter\def\csname cbBpDeclaredL2\endcsname{15}
\expandafter\def\csname cbBpDeclaredL2I\endcsname{20}
\expandafter\def\csname cbBpFortyBoth\endcsname{88}
\expandafter\def\csname cbBpFortyNeither\endcsname{43}
\expandafter\def\csname cbBpFortyUnrealOnly\endcsname{22}
\expandafter\def\csname cbBpFortyVendorOnly\endcsname{22}
\expandafter\def\csname cbBpRawBoth\endcsname{99}
\expandafter\def\csname cbBpRawNeither\endcsname{33}
\expandafter\def\csname cbBpRawUnrealOnly\endcsname{21}
\expandafter\def\csname cbBpRawVendorOnly\endcsname{22}
\expandafter\def\csname cbBpSensitivityN\endcsname{56}
\expandafter\def\csname cbBpShortCeiling\endcsname{0}
\expandafter\def\csname cbBpUnrealAnyMcp\endcsname{174}
\expandafter\def\csname cbBpUnrealDispatch\endcsname{173}
\expandafter\def\csname cbBpUnrealForty\endcsname{110}
\expandafter\def\csname cbBpUnrealFortyPct\endcsname{62.9}
\expandafter\def\csname cbBpUnrealGids\endcsname{0}
\expandafter\def\csname cbBpUnrealN\endcsname{175}
\expandafter\def\csname cbBpUnrealPostPermission\endcsname{4}
\expandafter\def\csname cbBpUnrealPrePermission\endcsname{171}
\expandafter\def\csname cbBpUnrealRaw\endcsname{120}
\expandafter\def\csname cbBpUnrealRawPct\endcsname{68.6}
\expandafter\def\csname cbBpUnrealRuntimeFailureSensitivity\endcsname{16}
\expandafter\def\csname cbBpUnrealRuntimeFailureTasks\endcsname{9}
\expandafter\def\csname cbBpUnrealRuntimeOnlyFailure\endcsname{25}
\expandafter\def\csname cbBpUnrealStructOnlyFailure\endcsname{1}
\expandafter\def\csname cbBpUnrealTranscript\endcsname{0}
\expandafter\def\csname cbBpVendorAnyMcp\endcsname{170}
\expandafter\def\csname cbBpVendorDispatch\endcsname{0}
\expandafter\def\csname cbBpVendorForty\endcsname{110}
\expandafter\def\csname cbBpVendorFortyPct\endcsname{62.9}
\expandafter\def\csname cbBpVendorGids\endcsname{0}
\expandafter\def\csname cbBpVendorN\endcsname{175}
\expandafter\def\csname cbBpVendorPostPermission\endcsname{7}
\expandafter\def\csname cbBpVendorPrePermission\endcsname{168}
\expandafter\def\csname cbBpVendorRaw\endcsname{121}
\expandafter\def\csname cbBpVendorRawPct\endcsname{69.1}
\expandafter\def\csname cbBpVendorRuntimeFailureSensitivity\endcsname{17}
\expandafter\def\csname cbBpVendorRuntimeFailureTasks\endcsname{7}
\expandafter\def\csname cbBpVendorRuntimeOnlyFailure\endcsname{28}
\expandafter\def\csname cbBpVendorStructOnlyFailure\endcsname{0}
\expandafter\def\csname cbBpVendorTranscript\endcsname{0}
\expandafter\def\csname cbContextAllLarger\endcsname{185}
\expandafter\def\csname cbContextDelta\endcsname{82,354}
\expandafter\def\csname cbContextN\endcsname{185}
\expandafter\def\csname cbContextRatio\endcsname{3.99}
\expandafter\def\csname cbContextSameDriverN\endcsname{0}
\expandafter\def\csname cbContextSameDriverRatio\endcsname{NA}
\expandafter\def\csname cbContextUnreal\endcsname{26,512}
\expandafter\def\csname cbContextUnusedDelta\endcsname{82,733}
\expandafter\def\csname cbContextUnusedN\endcsname{110}
\expandafter\def\csname cbContextVendor\endcsname{110,301}
\expandafter\def\csname cbCppConfigN\endcsname{693}
\expandafter\def\csname cbCppConfigprovidersserved\endcsname{561}
\expandafter\def\csname cbCppConfigreasoningpolicy\endcsname{693}
\expandafter\def\csname cbCppConfigsystempromptsha\endcsname{0}
\expandafter\def\csname cbCppConfigtoolsoffered\endcsname{0}
\expandafter\def\csname cbCppControlAnyMcp\endcsname{0}
\expandafter\def\csname cbCppControlDispatch\endcsname{0}
\expandafter\def\csname cbCppControlForty\endcsname{180}
\expandafter\def\csname cbCppControlFortyPct\endcsname{77.9}
\expandafter\def\csname cbCppControlGids\endcsname{165}
\expandafter\def\csname cbCppControlN\endcsname{231}
\expandafter\def\csname cbCppControlPostPermission\endcsname{181}
\expandafter\def\csname cbCppControlPrePermission\endcsname{50}
\expandafter\def\csname cbCppControlRaw\endcsname{181}
\expandafter\def\csname cbCppControlRawPct\endcsname{78.4}
\expandafter\def\csname cbCppControlTranscript\endcsname{50}
\expandafter\def\csname cbCppDeclaredL1\endcsname{33}
\expandafter\def\csname cbCppDeclaredL2\endcsname{33}
\expandafter\def\csname cbCppDeclaredL2I\endcsname{0}
\expandafter\def\csname cbCppFortyBoth\endcsname{188}
\expandafter\def\csname cbCppFortyNeither\endcsname{17}
\expandafter\def\csname cbCppFortyUnrealOnly\endcsname{11}
\expandafter\def\csname cbCppFortyVendorOnly\endcsname{15}
\expandafter\def\csname cbCppRawBoth\endcsname{192}
\expandafter\def\csname cbCppRawNeither\endcsname{15}
\expandafter\def\csname cbCppRawUnrealOnly\endcsname{10}
\expandafter\def\csname cbCppRawVendorOnly\endcsname{14}
\expandafter\def\csname cbCppShortCeiling\endcsname{1}
\expandafter\def\csname cbCppTranscripts\endcsname{512}
\expandafter\def\csname cbCppUnrealAnyMcp\endcsname{37}
\expandafter\def\csname cbCppUnrealDispatch\endcsname{9}
\expandafter\def\csname cbCppUnrealForty\endcsname{199}
\expandafter\def\csname cbCppUnrealFortyPct\endcsname{86.1}
\expandafter\def\csname cbCppUnrealGids\endcsname{198}
\expandafter\def\csname cbCppUnrealN\endcsname{231}
\expandafter\def\csname cbCppUnrealPostPermission\endcsname{0}
\expandafter\def\csname cbCppUnrealPrePermission\endcsname{231}
\expandafter\def\csname cbCppUnrealRaw\endcsname{202}
\expandafter\def\csname cbCppUnrealRawPct\endcsname{87.4}
\expandafter\def\csname cbCppUnrealTranscript\endcsname{231}
\expandafter\def\csname cbCppVendorAnyMcp\endcsname{95}
\expandafter\def\csname cbCppVendorDispatch\endcsname{0}
\expandafter\def\csname cbCppVendorForty\endcsname{203}
\expandafter\def\csname cbCppVendorFortyPct\endcsname{87.9}
\expandafter\def\csname cbCppVendorGids\endcsname{198}
\expandafter\def\csname cbCppVendorN\endcsname{231}
\expandafter\def\csname cbCppVendorPostPermission\endcsname{231}
\expandafter\def\csname cbCppVendorPrePermission\endcsname{0}
\expandafter\def\csname cbCppVendorRaw\endcsname{206}
\expandafter\def\csname cbCppVendorRawPct\endcsname{89.2}
\expandafter\def\csname cbCppVendorTranscript\endcsname{231}
\expandafter\def\csname cbMatchedN\endcsname{70}
\expandafter\def\csname cbMatchedUnrealBp\endcsname{32}
\expandafter\def\csname cbMatchedUnrealBpForty\endcsname{25}
\expandafter\def\csname cbMatchedUnrealCpp\endcsname{49}
\expandafter\def\csname cbMatchedUnrealCppForty\endcsname{46}
\expandafter\def\csname cbMatchedVendorBp\endcsname{30}
\expandafter\def\csname cbMatchedVendorBpForty\endcsname{20}
\expandafter\def\csname cbMatchedVendorCpp\endcsname{52}
\expandafter\def\csname cbMatchedVendorCppForty\endcsname{50}
\expandafter\def\csname cbModel0BpUnrealForty\endcsname{19}
\expandafter\def\csname cbModel0BpUnrealRaw\endcsname{20}
\expandafter\def\csname cbModel0BpVendorForty\endcsname{17}
\expandafter\def\csname cbModel0BpVendorRaw\endcsname{18}
\expandafter\def\csname cbModel0CppControlForty\endcsname{27}
\expandafter\def\csname cbModel0CppControlRaw\endcsname{27}
\expandafter\def\csname cbModel0CppUnrealForty\endcsname{30}
\expandafter\def\csname cbModel0CppUnrealRaw\endcsname{30}
\expandafter\def\csname cbModel0CppVendorForty\endcsname{30}
\expandafter\def\csname cbModel0CppVendorRaw\endcsname{30}
\expandafter\def\csname cbModel0PythonUnrealForty\endcsname{12}
\expandafter\def\csname cbModel0PythonUnrealRaw\endcsname{12}
\expandafter\def\csname cbModel0PythonVendorForty\endcsname{9}
\expandafter\def\csname cbModel0PythonVendorRaw\endcsname{9}
\expandafter\def\csname cbModel1BpUnrealForty\endcsname{17}
\expandafter\def\csname cbModel1BpUnrealRaw\endcsname{17}
\expandafter\def\csname cbModel1BpVendorForty\endcsname{15}
\expandafter\def\csname cbModel1BpVendorRaw\endcsname{19}
\expandafter\def\csname cbModel1CppControlForty\endcsname{26}
\expandafter\def\csname cbModel1CppControlRaw\endcsname{26}
\expandafter\def\csname cbModel1CppUnrealForty\endcsname{28}
\expandafter\def\csname cbModel1CppUnrealRaw\endcsname{28}
\expandafter\def\csname cbModel1CppVendorForty\endcsname{27}
\expandafter\def\csname cbModel1CppVendorRaw\endcsname{30}
\expandafter\def\csname cbModel1PythonUnrealForty\endcsname{12}
\expandafter\def\csname cbModel1PythonUnrealRaw\endcsname{12}
\expandafter\def\csname cbModel1PythonVendorForty\endcsname{11}
\expandafter\def\csname cbModel1PythonVendorRaw\endcsname{11}
\expandafter\def\csname cbModel2BpUnrealForty\endcsname{15}
\expandafter\def\csname cbModel2BpUnrealRaw\endcsname{15}
\expandafter\def\csname cbModel2BpVendorForty\endcsname{17}
\expandafter\def\csname cbModel2BpVendorRaw\endcsname{17}
\expandafter\def\csname cbModel2CppControlForty\endcsname{23}
\expandafter\def\csname cbModel2CppControlRaw\endcsname{23}
\expandafter\def\csname cbModel2CppUnrealForty\endcsname{29}
\expandafter\def\csname cbModel2CppUnrealRaw\endcsname{29}
\expandafter\def\csname cbModel2CppVendorForty\endcsname{31}
\expandafter\def\csname cbModel2CppVendorRaw\endcsname{31}
\expandafter\def\csname cbModel2PythonUnrealForty\endcsname{8}
\expandafter\def\csname cbModel2PythonUnrealRaw\endcsname{8}
\expandafter\def\csname cbModel2PythonVendorForty\endcsname{8}
\expandafter\def\csname cbModel2PythonVendorRaw\endcsname{8}
\expandafter\def\csname cbModel3BpUnrealForty\endcsname{13}
\expandafter\def\csname cbModel3BpUnrealRaw\endcsname{13}
\expandafter\def\csname cbModel3BpVendorForty\endcsname{16}
\expandafter\def\csname cbModel3BpVendorRaw\endcsname{16}
\expandafter\def\csname cbModel3CppControlForty\endcsname{21}
\expandafter\def\csname cbModel3CppControlRaw\endcsname{21}
\expandafter\def\csname cbModel3CppUnrealForty\endcsname{29}
\expandafter\def\csname cbModel3CppUnrealRaw\endcsname{29}
\expandafter\def\csname cbModel3CppVendorForty\endcsname{27}
\expandafter\def\csname cbModel3CppVendorRaw\endcsname{27}
\expandafter\def\csname cbModel3PythonUnrealForty\endcsname{6}
\expandafter\def\csname cbModel3PythonUnrealRaw\endcsname{6}
\expandafter\def\csname cbModel3PythonVendorForty\endcsname{10}
\expandafter\def\csname cbModel3PythonVendorRaw\endcsname{10}
\expandafter\def\csname cbModel4BpUnrealForty\endcsname{18}
\expandafter\def\csname cbModel4BpUnrealRaw\endcsname{18}
\expandafter\def\csname cbModel4BpVendorForty\endcsname{19}
\expandafter\def\csname cbModel4BpVendorRaw\endcsname{19}
\expandafter\def\csname cbModel4CppControlForty\endcsname{29}
\expandafter\def\csname cbModel4CppControlRaw\endcsname{29}
\expandafter\def\csname cbModel4CppUnrealForty\endcsname{30}
\expandafter\def\csname cbModel4CppUnrealRaw\endcsname{30}
\expandafter\def\csname cbModel4CppVendorForty\endcsname{30}
\expandafter\def\csname cbModel4CppVendorRaw\endcsname{30}
\expandafter\def\csname cbModel4PythonUnrealForty\endcsname{9}
\expandafter\def\csname cbModel4PythonUnrealRaw\endcsname{9}
\expandafter\def\csname cbModel4PythonVendorForty\endcsname{8}
\expandafter\def\csname cbModel4PythonVendorRaw\endcsname{8}
\expandafter\def\csname cbModel5BpUnrealForty\endcsname{17}
\expandafter\def\csname cbModel5BpUnrealRaw\endcsname{19}
\expandafter\def\csname cbModel5BpVendorForty\endcsname{18}
\expandafter\def\csname cbModel5BpVendorRaw\endcsname{20}
\expandafter\def\csname cbModel5CppControlForty\endcsname{28}
\expandafter\def\csname cbModel5CppControlRaw\endcsname{28}
\expandafter\def\csname cbModel5CppUnrealForty\endcsname{30}
\expandafter\def\csname cbModel5CppUnrealRaw\endcsname{30}
\expandafter\def\csname cbModel5CppVendorForty\endcsname{30}
\expandafter\def\csname cbModel5CppVendorRaw\endcsname{30}
\expandafter\def\csname cbModel5PythonUnrealForty\endcsname{10}
\expandafter\def\csname cbModel5PythonUnrealRaw\endcsname{11}
\expandafter\def\csname cbModel5PythonVendorForty\endcsname{11}
\expandafter\def\csname cbModel5PythonVendorRaw\endcsname{11}
\expandafter\def\csname cbModel6BpUnrealForty\endcsname{11}
\expandafter\def\csname cbModel6BpUnrealRaw\endcsname{18}
\expandafter\def\csname cbModel6BpVendorForty\endcsname{8}
\expandafter\def\csname cbModel6BpVendorRaw\endcsname{12}
\expandafter\def\csname cbModel6CppControlForty\endcsname{26}
\expandafter\def\csname cbModel6CppControlRaw\endcsname{27}
\expandafter\def\csname cbModel6CppUnrealForty\endcsname{23}
\expandafter\def\csname cbModel6CppUnrealRaw\endcsname{26}
\expandafter\def\csname cbModel6CppVendorForty\endcsname{28}
\expandafter\def\csname cbModel6CppVendorRaw\endcsname{28}
\expandafter\def\csname cbModel6PythonUnrealForty\endcsname{7}
\expandafter\def\csname cbModel6PythonUnrealRaw\endcsname{10}
\expandafter\def\csname cbModel6PythonVendorForty\endcsname{4}
\expandafter\def\csname cbModel6PythonVendorRaw\endcsname{5}
\expandafter\def\csname cbPythonConfigN\endcsname{168}
\expandafter\def\csname cbPythonConfigprovidersserved\endcsname{0}
\expandafter\def\csname cbPythonConfigreasoningpolicy\endcsname{149}
\expandafter\def\csname cbPythonConfigsystempromptsha\endcsname{0}
\expandafter\def\csname cbPythonConfigtoolsoffered\endcsname{0}
\expandafter\def\csname cbPythonDeclaredL1\endcsname{12}
\expandafter\def\csname cbPythonDeclaredL2\endcsname{0}
\expandafter\def\csname cbPythonDeclaredL2I\endcsname{12}
\expandafter\def\csname cbPythonFortyBoth\endcsname{48}
\expandafter\def\csname cbPythonFortyNeither\endcsname{7}
\expandafter\def\csname cbPythonFortyUnrealOnly\endcsname{16}
\expandafter\def\csname cbPythonFortyVendorOnly\endcsname{13}
\expandafter\def\csname cbPythonRawBoth\endcsname{52}
\expandafter\def\csname cbPythonRawNeither\endcsname{6}
\expandafter\def\csname cbPythonRawUnrealOnly\endcsname{16}
\expandafter\def\csname cbPythonRawVendorOnly\endcsname{10}
\expandafter\def\csname cbPythonShortCeiling\endcsname{0}
\expandafter\def\csname cbPythonUnrealAnyMcp\endcsname{71}
\expandafter\def\csname cbPythonUnrealDispatch\endcsname{70}
\expandafter\def\csname cbPythonUnrealForty\endcsname{64}
\expandafter\def\csname cbPythonUnrealFortyPct\endcsname{76.2}
\expandafter\def\csname cbPythonUnrealGids\endcsname{0}
\expandafter\def\csname cbPythonUnrealN\endcsname{84}
\expandafter\def\csname cbPythonUnrealPostPermission\endcsname{41}
\expandafter\def\csname cbPythonUnrealPrePermission\endcsname{43}
\expandafter\def\csname cbPythonUnrealRaw\endcsname{68}
\expandafter\def\csname cbPythonUnrealRawPct\endcsname{81.0}
\expandafter\def\csname cbPythonUnrealTranscript\endcsname{0}
\expandafter\def\csname cbPythonVendorAnyMcp\endcsname{73}
\expandafter\def\csname cbPythonVendorDispatch\endcsname{0}
\expandafter\def\csname cbPythonVendorForty\endcsname{61}
\expandafter\def\csname cbPythonVendorFortyPct\endcsname{72.6}
\expandafter\def\csname cbPythonVendorGids\endcsname{0}
\expandafter\def\csname cbPythonVendorN\endcsname{84}
\expandafter\def\csname cbPythonVendorPostPermission\endcsname{42}
\expandafter\def\csname cbPythonVendorPrePermission\endcsname{42}
\expandafter\def\csname cbPythonVendorRaw\endcsname{62}
\expandafter\def\csname cbPythonVendorRawPct\endcsname{73.8}
\expandafter\def\csname cbPythonVendorTranscript\endcsname{0}
\expandafter\def\csname cbSnapshot\endcsname{cc848c18ca50}
\expandafter\def\csname cbTotal\endcsname{1211}

\newcommand{\trajvalue}[1]{\csname tr#1\endcsname}
\expandafter\def\csname trAfterDirectiveRuns\endcsname{33}
\expandafter\def\csname trAllTranscripts\endcsname{1211}
\expandafter\def\csname trAuraBpOnly\endcsname{2}
\expandafter\def\csname trAuraCppOnly\endcsname{24}
\expandafter\def\csname trAuraExplicitBpBehaviorFailures\endcsname{28}
\expandafter\def\csname trAuraRuntimeAfterStructure\endcsname{14}
\expandafter\def\csname trAuraStructurePassLtwoFail\endcsname{28}
\expandafter\def\csname trBpAuraOnly\endcsname{22}
\expandafter\def\csname trBpBoth\endcsname{99}
\expandafter\def\csname trBpCallRatio\endcsname{0.50}
\expandafter\def\csname trBpCallRatioHigh\endcsname{0.57}
\expandafter\def\csname trBpCallRatioLow\endcsname{0.45}
\expandafter\def\csname trBpDurationRatio\endcsname{1.25}
\expandafter\def\csname trBpDurationRatioHigh\endcsname{1.49}
\expandafter\def\csname trBpDurationRatioLow\endcsname{1.09}
\expandafter\def\csname trBpNeither\endcsname{33}
\expandafter\def\csname trBpPairs\endcsname{175}
\expandafter\def\csname trBpPromptConnected\endcsname{166}
\expandafter\def\csname trBpPromptConnectedJudge\endcsname{149}
\expandafter\def\csname trBpPromptMatched\endcsname{171}
\expandafter\def\csname trBpUnrealOnly\endcsname{21}
\expandafter\def\csname trDelegatedBpRuns\endcsname{109}
\expandafter\def\csname trDirectiveCalls\endcsname{45}
\expandafter\def\csname trDirectiveRuns\endcsname{38}
\expandafter\def\csname trDslAcceptedRuns\endcsname{55}
\expandafter\def\csname trDslCalls\endcsname{454}
\expandafter\def\csname trDslErrorCalls\endcsname{189}
\expandafter\def\csname trDslErrorRuns\endcsname{58}
\expandafter\def\csname trDslReadCompileFail\endcsname{14}
\expandafter\def\csname trDslReadCompileRuns\endcsname{42}
\expandafter\def\csname trDslReadCompileRuntimeFail\endcsname{9}
\expandafter\def\csname trNoTerminalSummary\endcsname{34}
\expandafter\def\csname trOutcomeBootstrap\endcsname{50000}
\expandafter\def\csname trPairedRuns\endcsname{490}
\expandafter\def\csname trPythonStartupFailures\endcsname{11}
\expandafter\def\csname trPythonStartupKeptAuraN\endcsname{73}
\expandafter\def\csname trPythonStartupKeptAuraPass\endcsname{62}
\expandafter\def\csname trPythonStartupKeptUnrealN\endcsname{73}
\expandafter\def\csname trPythonStartupKeptUnrealPass\endcsname{59}
\expandafter\def\csname trSameScriptFail\endcsname{2}
\expandafter\def\csname trSameScriptRuns\endcsname{18}
\expandafter\def\csname trSurfacePairs\endcsname{140}
\expandafter\def\csname trUnrealBpOnly\endcsname{3}
\expandafter\def\csname trUnrealCppOnly\endcsname{20}
\expandafter\def\csname trUnrealExplicitBpBehaviorFailures\endcsname{24}
\expandafter\def\csname trUnrealRuntimeAfterStructure\endcsname{13}
\expandafter\def\csname trUnrealStructurePassLtwoFail\endcsname{25}
\expandafter\def\csname trWorkBootstrap\endcsname{20000}
\expandafter\def\csname cbCppTranscripts\endcsname{693}
\expandafter\def\csname cbCppControlTranscript\endcsname{231}
\expandafter\def\csname cbCppUnrealTranscript\endcsname{231}
\expandafter\def\csname cbCppVendorTranscript\endcsname{231}
\expandafter\def\csname cbBpTranscripts\endcsname{350}
\expandafter\def\csname cbBpUnrealTranscript\endcsname{175}
\expandafter\def\csname cbBpVendorTranscript\endcsname{175}
\expandafter\def\csname cbPythonTranscripts\endcsname{168}
\expandafter\def\csname cbPythonUnrealTranscript\endcsname{84}
\expandafter\def\csname cbPythonVendorTranscript\endcsname{84}

\newcommand{\framevalue}[1]{\csname fr#1\endcsname}
\expandafter\def\csname frAuraMinGapDropTwo\endcsname{16.1}
\expandafter\def\csname frBpAuraCalls\endcsname{9016}
\expandafter\def\csname frBpAuraFullyTimedRuns\endcsname{171}
\expandafter\def\csname frBpAuraTimedCalls\endcsname{9012}
\expandafter\def\csname frBpDelegateCalls\endcsname{200}
\expandafter\def\csname frBpDelegateCompleteShare\endcsname{92.4}
\expandafter\def\csname frBpDelegateCompleteUsers\endcsname{106}
\expandafter\def\csname frBpDelegateTimedCalls\endcsname{197}
\expandafter\def\csname frBpDelegateUsers\endcsname{109}
\expandafter\def\csname frBpDelegateWaitShare\endcsname{92.1}
\expandafter\def\csname frBpFullyTimedPairN\endcsname{171}
\expandafter\def\csname frBpFullyTimedWaitRatio\endcsname{9.24}
\expandafter\def\csname frBpMissingDelegateReturns\endcsname{3}
\expandafter\def\csname frBpMissingReturns\endcsname{4}
\expandafter\def\csname frBpUnrealCalls\endcsname{16929}
\expandafter\def\csname frBpUnrealFullyTimedRuns\endcsname{175}
\expandafter\def\csname frBpUnrealTimedCalls\endcsname{16929}
\expandafter\def\csname frBpWaitRatio\endcsname{8.36}
\expandafter\def\csname frBpWaitRatioHigh\endcsname{13.83}
\expandafter\def\csname frBpWaitRatioLow\endcsname{5.35}
\expandafter\def\csname frDropTwoCombinations\endcsname{45}
\expandafter\def\csname frInitFailureCases\endcsname{7}
\expandafter\def\csname frInitOtherUnresolvedCases\endcsname{2}
\expandafter\def\csname frInitTableCases\endcsname{4}
\expandafter\def\csname frUnrealMinGapDropTwo\endcsname{17.9}

\newcommand{\dispatchvalue}[1]{\csname sd#1\endcsname}
\expandafter\def\csname sdUnrealBpEngaged\endcsname{173}
\expandafter\def\csname sdUnrealBpEngagedPct\endcsname{98.9}
\expandafter\def\csname sdUnrealBpMedianOps\endcsname{48}
\expandafter\def\csname sdUnrealBpN\endcsname{175}
\expandafter\def\csname sdUnrealCppEngaged\endcsname{9}
\expandafter\def\csname sdUnrealCppEngagedPct\endcsname{3.9}
\expandafter\def\csname sdUnrealCppMedianOps\endcsname{0}
\expandafter\def\csname sdUnrealCppN\endcsname{231}
\expandafter\def\csname sdUnrealPythonEngaged\endcsname{70}
\expandafter\def\csname sdUnrealPythonEngagedPct\endcsname{83.3}
\expandafter\def\csname sdUnrealPythonMedianOps\endcsname{25}
\expandafter\def\csname sdUnrealPythonN\endcsname{84}
\expandafter\def\csname sdUnrealSurfacePanelModels\endcsname{7}
\expandafter\def\csname sdUnrealSurfaceUnanimousModels\endcsname{7}

\newcommand{\studyvalue}[1]{\csname study#1\endcsname}
\expandafter\def\csname studyAuraBp\endcsname{20}
\expandafter\def\csname studyAuraBpOnly\endcsname{2}
\expandafter\def\csname studyAuraCpp\endcsname{50}
\expandafter\def\csname studyAuraCppOnly\endcsname{32}
\expandafter\def\csname studyAuraGap\endcsname{42.9}
\expandafter\def\csname studyAuraLateBpInCppOnly\endcsname{8}
\expandafter\def\csname studyAuraMinDropTwo\endcsname{30.4}
\expandafter\def\csname studyAuraRuntimeAfterStructure\endcsname{14}
\expandafter\def\csname studyBpAuraN\endcsname{175}
\expandafter\def\csname studyBpAuraPass\endcsname{110}
\expandafter\def\csname studyBpAuraPct\endcsname{62.9}
\expandafter\def\csname studyBpUnrealN\endcsname{175}
\expandafter\def\csname studyBpUnrealPass\endcsname{110}
\expandafter\def\csname studyBpUnrealPct\endcsname{62.9}
\expandafter\def\csname studyCppAuraN\endcsname{231}
\expandafter\def\csname studyCppAuraPass\endcsname{203}
\expandafter\def\csname studyCppAuraPct\endcsname{87.9}
\expandafter\def\csname studyCppControlN\endcsname{231}
\expandafter\def\csname studyCppControlPass\endcsname{180}
\expandafter\def\csname studyCppControlPct\endcsname{77.9}
\expandafter\def\csname studyCppUnrealN\endcsname{231}
\expandafter\def\csname studyCppUnrealPass\endcsname{199}
\expandafter\def\csname studyCppUnrealPct\endcsname{86.1}
\expandafter\def\csname studyPythonAuraN\endcsname{84}
\expandafter\def\csname studyPythonAuraPass\endcsname{66}
\expandafter\def\csname studyPythonAuraPct\endcsname{78.6}
\expandafter\def\csname studyPythonUnrealN\endcsname{84}
\expandafter\def\csname studyPythonUnrealPass\endcsname{62}
\expandafter\def\csname studyPythonUnrealPct\endcsname{73.8}
\expandafter\def\csname studyUnrealBp\endcsname{25}
\expandafter\def\csname studyUnrealBpOnly\endcsname{2}
\expandafter\def\csname studyUnrealCpp\endcsname{46}
\expandafter\def\csname studyUnrealCppOnly\endcsname{23}
\expandafter\def\csname studyUnrealGap\endcsname{30.0}
\expandafter\def\csname studyUnrealLateBpInCppOnly\endcsname{5}
\expandafter\def\csname studyUnrealMinDropTwo\endcsname{21.4}
\expandafter\def\csname studyUnrealRuntimeAfterStructure\endcsname{12}

\newcommand{\nummatchedanalysis}{10}
\newcommand{\matchedanalysisunrealgap}{30.0}
\newcommand{\matchedanalysismcpgap}{42.9}

\newcommand{\revisionvalue}[1]{\csname revisionvalue#1\endcsname}
\expandafter\def\csname revisionvalueGameplayRuns\endcsname{70}
\expandafter\def\csname revisionvalueUnrealAssetEligible\endcsname{45}
\expandafter\def\csname revisionvalueUnrealRuntimeFail\endcsname{19}
\expandafter\def\csname revisionvalueUnrealRuntimeFailPct\endcsname{42.2}
\expandafter\def\csname revisionvalueUnrealStateReadback\endcsname{15}
\expandafter\def\csname revisionvalueUnrealAssetReadback\endcsname{62}
\expandafter\def\csname revisionvalueUnrealCppPass\endcsname{46}
\expandafter\def\csname revisionvalueUnrealBpPass\endcsname{25}
\expandafter\def\csname revisionvalueUnrealGap\endcsname{30.0}
\expandafter\def\csname revisionvalueMcpBAssetEligible\endcsname{40}
\expandafter\def\csname revisionvalueMcpBRuntimeFail\endcsname{20}
\expandafter\def\csname revisionvalueMcpBRuntimeFailPct\endcsname{50.0}
\expandafter\def\csname revisionvalueMcpBStateReadback\endcsname{19}
\expandafter\def\csname revisionvalueMcpBAssetReadback\endcsname{69}
\expandafter\def\csname revisionvalueMcpBCppPass\endcsname{50}
\expandafter\def\csname revisionvalueMcpBBpPass\endcsname{20}
\expandafter\def\csname revisionvalueMcpBGap\endcsname{42.9}


\ifanon\else
  \iclrfinalcopy
\fi

\newcommand{\blfootnote}[1]{%
  \begingroup\renewcommand{\thefootnote}{}\footnote{#1}%
  \addtocounter{footnote}{-1}\endgroup}

\newcommand{\papertitle}{\benchname{}: Deterministic Evaluation for Coding Agents in Unreal Engine}

\title{\papertitle}
\ifanon
\author{Anonymous authors\\Paper under double-blind review}
\fi

\begin{document}
\ifanon
\maketitle
\else
% --- arXiv title block (GameEngineBench-style: title, rule, authors grouped
% --- by affiliation, contact line). The September ICLR build ignores this
% --- branch and uses the style file's \maketitle.
\begin{flushleft}
{\LARGE\bfseries\raggedright \papertitle{\normalfont\normalsize}\par}
\vspace{0.7em}
{\color{lightgray}\hrule height 0.5pt}
\vspace{1.5em}
{Shutong Wu\textsuperscript{*}}\\
{\small University of Rochester}\\[0.9em]
{Kevin Calderone, Andy Tsen\textsuperscript{\dag}}\\
{\small RamenVR}\\[1.4em]
{\small \faEnvelope\;\texttt{swu85@ur.rochester.edu} \qquad
\faGithub\;\artifacturl}
\end{flushleft}
\blfootnote{\textsuperscript{*}Work done during internship at RamenVR.}
\blfootnote{\textsuperscript{\dag}Corresponding author.}
\vspace{1.8em}
\fi

% Redline key for the ICLR root. \revisionlegend self-suppresses when
% ACCEPTREVISIONS is defined; measured, it costs no page in either build.
\begin{abstract}
Building gameplay features in a game engine requires more than code, as code that compiles and runs does not necessarily implement the requested gameplay. We introduce \benchname{}, an evaluation harness that runs agents in an isolated Unreal Engine environment, reconstructs their saved submissions in fresh projects, and applies deterministic build, asset, and runtime checks without an LLM judge. Based on the harness, we build a benchmark consisting of \numtasks{} tasks spanning C++ source, Blueprint assets, and editor scripting. We evaluate \numpanelmodels{} models under two editor-tool configurations, with a file-and-shell baseline on C++ tasks. We further pair tasks that specify the same gameplay and use the same runtime tests, but require C++ and Blueprint as the deliverables. Across the \nummatchedanalysis{} paired tasks, C++ completion rates exceed Blueprint by \matchedanalysisunrealgap{} and \matchedanalysismcpgap{} percentage points in the two tool configurations. Among on-time Blueprint submissions in this paired set that pass asset checks, \revisionvalue{UnrealRuntimeFailPct}\%  and \revisionvalue{McpBRuntimeFailPct}\% fail explicit runtime assertions. These submissions satisfy asset requirements but fail the required gameplay tests. We will release the harness, task benchmark, and our trajectory findings with the report.
\end{abstract}

% Show representative tasks early in the public preprint.
\ifanon\else
\begin{figure}[!ht]
\centering
\includegraphics[width=\linewidth]{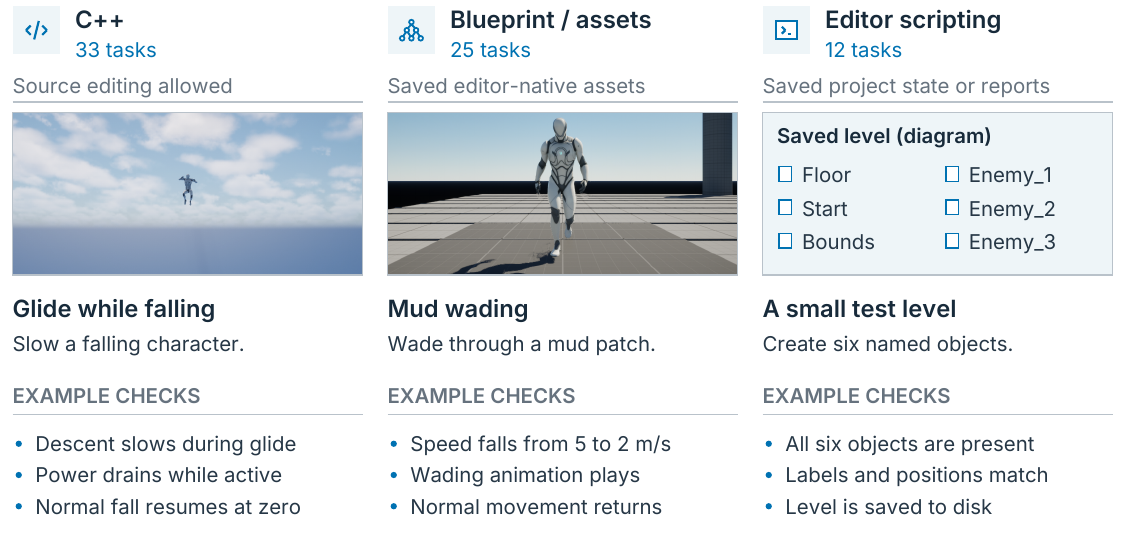}
\caption{Sample benchmark tasks across the three deliverable groups. The screenshots illustrate the glide task (\texttt{gp-glide-stamina-cpp}) and mud-wading task (\texttt{t1-mud-wade-bp}). The editor-scripting panel shows the \texttt{kp-spawn-level-actors} task, which requires creating and saving a level containing six named objects.}
\label{fig:task-overview}
\end{figure}

\fi
\section{Introduction}
\label{sec:intro}

Implementing a gameplay feature involves more than producing code that compiles. In Unreal Engine, a feature can span C++ classes, Blueprint graphs, and other features such as the Niagara VFX system. Blueprint is Unreal's visual programming system, which uses node graphs saved in binary \texttt{.uasset} files. The evaluation question is therefore whether the saved project implements the requested behavior, not simply whether it compiles.

Existing benchmarks evaluate generated functions, repository changes, and tasks in interactive environments~\citep{humaneval,swebench,zhou2024webarena,terminalbench}. Game-development benchmarks extend executable evaluation to game engines such as Godot, Unreal, and Roblox~\citep{gamedevbench,gameenginebench,opengameeval}. Evaluating projects that combine source code and native assets requires checking both the submitted files and the behavior they produce. The benchmark harness should also differentiate between structural requirements, like a valid asset reference or successful compilation, and behavioral requirements, such as a character decelerating while gliding.

We introduce \benchname{}, an evaluation harness for coding agents working with Unreal projects. Each task defines the starting state and the goal deliverables. The harness collects the generated work, then reconstructs it in a fresh project and grades it based on the build, asset, and runtime checks. Runtime checks observe behavior in Play-in-Editor (\pie{}), Unreal's mode for running a game inside the editor.

Based on the harness, we construct a benchmark of \numtasks{} bounded tasks: \numcpp{} C++ tasks, \numbp{} Blueprint/native-asset tasks, and \numpython{} editor-scripting tasks. The task suite contains \nummatchedpairs{} gameplay task pairs, each with both C++ and Blueprint versions. This design enables analysis of how the same agent approaches the same tasks across different deliverable formats.

We evaluate \numpanelmodels{} models under two editor-tool configurations, \armmcp{} and \armaura{} (\S\ref{sec:setup}), with a file-and-shell C++ baseline. In the paired analysis, we found that C++ completion exceeds Blueprint completion by \matchedanalysisunrealgap{} and \matchedanalysismcpgap{} percentage points under \armmcp{} and \armaura{} within the time budget. Among on-time Blueprint submissions in this set that satisfy asset checks, \revisionvalue{UnrealRuntimeFailPct}\% (\revisionvalue{UnrealRuntimeFail}/\revisionvalue{UnrealAssetEligible}) under \armmcp{} and \revisionvalue{McpBRuntimeFailPct}\% (\revisionvalue{McpBRuntimeFail}/\revisionvalue{McpBAssetEligible}) under \armaura{} fail explicit runtime assertions. The comparison and result analyses address complementary questions: whether agents complete the same gameplay under different deliverable requirements, and what passing an asset check tells us about the final completed feature.

Our contributions are:
\begin{enumerate}
\item \textbf{An Unreal evaluation harness.} \benchname{} runs an agent on a task in an isolated Unreal project, reconstructs its submission and grades it through build, asset, and runtime verifiers.
\item \textbf{A benchmark of \numtasks{} tasks.} The tasks cover source code, native assets, and editor scripting, including paired gameplay specifications with the same runtime tests.
\item \textbf{An empirical study of how coding agents work in game development tasks.} We share our results on how different models work through the benchmark, and how they perform and verify their work during the benchmarking process. Paired comparisons quantify differences under two editor configurations; submission and trajectory analyses show where asset construction and authoring feedback fall short of completed gameplay.
\end{enumerate}

\section{Related Work}
\label{sec:related}

\paragraph{Execution-based agent evaluation.}
Tests can reject a valid solution if they impose requirements that the task description does not state. SWE-bench Verified addresses this problem through developer review, screening out unclear tasks and tests that reject valid solutions~\citep{swebenchverified}. The checks should match what the agent was asked to do, whether the task involves writing a function, modifying a repository, or working in a terminal or desktop environment~\citep{humaneval,swebench,terminalbench,xie2024osworld}. We document our validation process in \S\ref{sec:benchmark}.

\paragraph{Game-development benchmarks.}
GameDevBench uses deterministic tests for Godot scripts, scenes, and resources and examines how image and video feedback affect agent performance~\citep{gamedevbench}. In GameEngineBench, agents make C++ changes in open source Unreal projects, and the changes are examined by both build and runtime tests; an LLM judge then considers the test results and implementation work~\citep{gameenginebench}. OpenGameEval evaluates Roblox Studio tasks with executable tests~\citep{opengameeval}. GameXpert-Bench spans generation, repair, and multi-turn optimization of browser-native JavaScript games~\citep{gamexpertbench}. These benchmarks already evaluate the behavior of changes made by agents. GBQA instead scores autonomous defect discovery in running web games by the share of human-verified bugs found~\citep{gbqa}.

Other work in Unreal covers cinematic assets and complete games. CutsceneBench combines tool and asset checks for Level Sequence assets with evaluation of rendered videos~\citep{cutsceneagent}. AutoUE focuses on end-to-end Unreal game generation using an LLM judge as the grader~\citep{autoue}. Our tasks focus on individual gameplay features with outcomes that could be deterministically checked. The harness reconstructs each submission in a fresh Unreal project and applies build, asset, and runtime checks. Appendix~\ref{app:benchmark-comparison} provides a detailed comparison of these benchmarks.

\paragraph{Interfaces and tool use.}
MCP exposes application operations as tools~\citep{anthropic2024mcp}, and MCP benchmarks study tool selection and composition~\citep{luo2025mcpuniverse,wu2025mcpmark}. Orak likewise standardizes agent access to games through MCP, but its deliverable is a gameplay trajectory scored by game outcomes rather than a saved engine artifact~\citep{orak}. MCP-Unity connects MCP clients through a Python server to a C\# Unity Editor plugin~\citep{wu2025mcpunity}; Epic's in-editor Unreal MCP server exposes tools over local HTTP and dispatches calls on the game thread~\citep{epic2026unrealmcp}. These interfaces provide authoring tool sets; our harness evaluates saved submissions in fresh Unreal projects.

Agent-interface studies show that the surrounding configuration can affect coding outcomes~\citep{sweagent,stopcomparing}. SkillsBench compares matched conditions with and without curated procedural Skills using deterministic verifiers~\citep{skillsbench}, whereas we vary editor configurations and required deliverables. We therefore describe editor services, editing permissions, and execution settings alongside the model panel. The observed differences compare these configurations; they do not identify the effect of an individual tool operation.

\section{\benchname{}}
\label{sec:benchmark}

\subsection{Evaluation setting and task scope}

\benchname{} measures whether an agent can implement a bounded gameplay or
editor behavior in an existing Unreal project. The unit of evaluation is an \emph{atomic mechanism task}: an instruction specifies a bounded gameplay or editor behavior, and states which project files or assets the agent may change. Examples include attaching an object to a skeletal socket, enforcing a cooldown, persisting a Blueprint graph, and authoring a mini level (Figure~\ref{fig:task-overview}). Each task focuses on a small set of engine mechanisms and uses named checks to identify which tested conditions fail. Tasks may still couple subsystems, as a movement behavior can involve input, state, timing, and physics, but they do not combine unrelated features merely to increase difficulty. Each score records whether the submitted project implements that mechanism under the stated task conditions.

\subsection{\benchname{} evaluation harness}
\label{sec:harness}
Container-oriented frameworks such as Harbor package instructions, environments, and tests~\citep{harborframework}. Unreal evaluation is comparatively more complex since the framework needs to control a versioned Editor and \pie{} environment, while preserving binary assets and saved changes. A task package specifies the starting project, prompt, writable area, engine build, verifier, and reference implementation. The harness reconstructs the project and delivers the prompt through the selected agent configuration. Reference implementations and verifier sources are withheld during authoring; detected exposure excludes a run from scoring (Appendix~\ref{app:hiding}).

\begin{figure}[t]
\centering
\includegraphics[width=\linewidth]{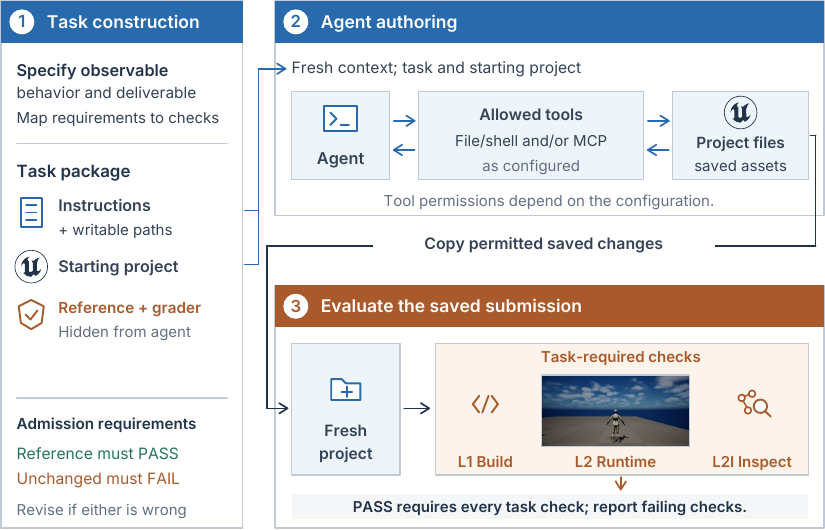}
\caption{Task construction, agent authoring, and evaluation of the saved submission in a fresh project. Each task declares its required checks. Reference-pass and unchanged-project-fail are admission criteria; \S\ref{sec:task-validation} describes task authoring and validation.}
\label{fig:task-lifecycle}
\end{figure}

The harness records interactions, timing, and project changes. When authoring ends, it collects the saved changes allowed by the task and applies them to a fresh copy of the starting project. This clean replay separates the submitted artifact from unrelated workspace modifications. It then builds the project, runs behavioral checks, and inspects native assets as declared by the task. Reports include the overall verdict, gate outcomes, and the failing checks identified. Figure~\ref{fig:task-lifecycle} summarizes the workflow.

\subsection{Execution-based deterministic grading}
\label{sec:grading}

Each task specifies which of the following checks its submission must pass:

\begin{enumerate}
\item
\textbf{Build (L1).} Compile the relevant editor and game targets. Submission-caused compiler or linker errors are failures.
\item
\textbf{Behavior (L2).} A fixed-timestep Play-in-Editor (\pie{}) session will start and evaluate named assertions at certain timesteps using Unreal Functional Tests.\footnote{Unreal's Functional Testing framework runs scripted checks inside a level; tests can be written in C++ or Blueprint. See \href{https://dev.epicgames.com/documentation/en-us/unreal-engine/functional-testing-in-unreal-engine}{Epic's Functional Testing documentation}.} Assertions focus on state transitions and variable changes during the session.
\item
\textbf{Artifact (L2I).} Load the saved assets in the editor without its graphical interface and check the required graph topology, class hierarchy, typed properties, references, or persistence.
\end{enumerate}

For each task, we specify the checks a submission must pass and evaluate them directly in Unreal Engine. The same assertions and checkpoint schedule are applied to every submission for a task. A fixed simulation timestep (1/60 second by default) makes time-dependent checks comparable, while per-task numerical tolerances accommodate the quantities being tested; Appendix~\ref{app:configuration} lists the engine switches and the tolerances in force. No LLM reads the submission, trace, screenshot, or test report to decide whether it passes or fails. Screenshots may be retained for diagnosis, but never to change the final grade. Structural checks establish required asset properties, while runtime checks establish the tested behavior. This excludes creative tasks whose success depends on visual judgment, a limitation we discuss in Sec.~\mbox{\ref{sec:limitations}}.

\subsection{The benchmark and required deliverables}
The task suite contains \numcpp{} C++ tasks, \numbp{} Blueprint tasks, and \numpython{} editor-scripting tasks. C++ tasks permit source code changes in a designated folder; BP tasks require saved native assets; scripting tasks are evaluated on their resulting saved project state. The tasks cover domains including gameplay logic, character movement, animation, UI, and editor operations. Of the \numtasks{} tasks, \numthirdperson{} start from the \ue{} Third Person template and \numtemplateproj{} from a customized template. Table~\ref{tab:verifier-coverage} gives their required checks; Appendix~\ref{app:catalog} lists the individual tasks.

Figure~\ref{fig:task-overview} shows the three task types in our benchmark. Tasks usually contain multiple checks: glide, for example, must slow a falling character, consume Power, and restore to normal falling state when Power is drained. Editor-scripting tasks instead check the saved output, such as whether a level contains the requested objects. The harness uses each task's declared checks since not all check types are needed for each task.

\ifanon
\begin{figure}[!ht]
\centering
\includegraphics[width=\linewidth]{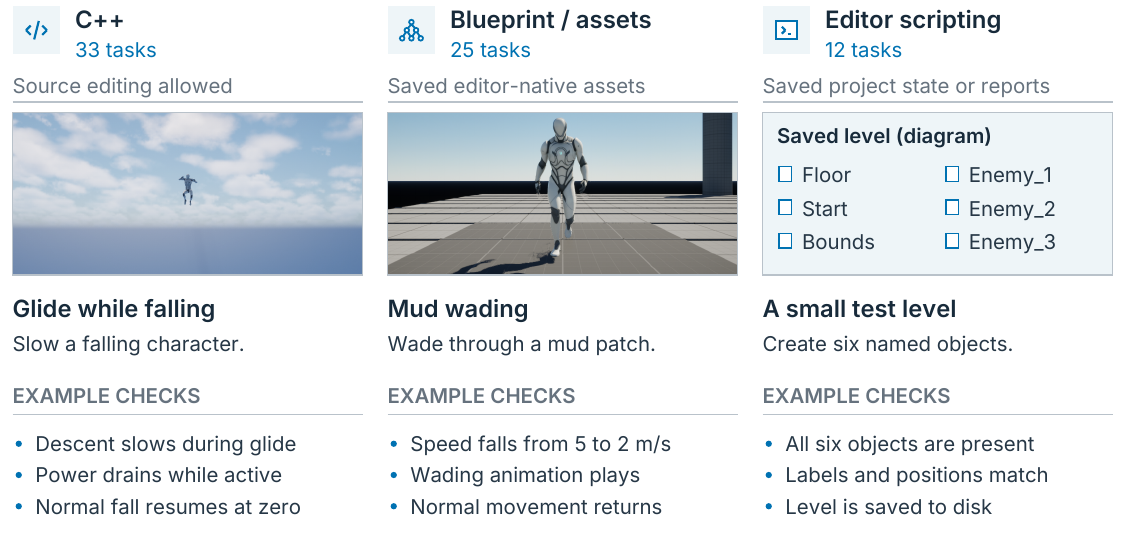}
\caption{Sample benchmark tasks across the three deliverable groups. The screenshots illustrate the glide task (\texttt{gp-glide-stamina-cpp}) and mud-wading task (\texttt{t1-mud-wade-bp}). The editor-scripting panel shows the \texttt{kp-spawn-level-actors} task, which requires creating and saving a level containing six named objects.}
\label{fig:task-overview}
\end{figure}

\fi

\begin{table}[!ht]
\centering\small
\begin{tabular}{@{}lrrrr@{}}
\toprule
Task group & Tasks & Build & Runtime & Assets \\
\midrule
C++ & \numcpp{} & \numcpp{} & \numcpp{} & 0 \\
BP & \numbp{} & \numbp{} & \auditvalue{BpDeclaredL2} & \auditvalue{BpDeclaredL2I} \\
Python & \numpython{} & \numpython{} & 0 & \numpython{} \\
\bottomrule
\end{tabular}
\caption{Required checks in the measured suite. Runtime and asset columns overlap; each paired BP gameplay task requires both.}
\label{tab:verifier-coverage}
\end{table}

Ten gameplay specifications have C++ and BP versions that share a starting project template and runtime verifiers. They contain the same interface and gameplay instruction; the only difference is the requested deliverable type (BP and C++). BP additionally requires saved assets. We inspect the submitted files when interpreting the comparison. \S\ref{sec:setup} defines the paired analysis.

\subsection{Task authoring and validation}
\label{sec:task-validation}
The tasks are designed around common Unreal workflows and informed by the team's engine expertise and public engine documentation. Team members, including artists, designers, and engineers, propose behaviors that represent the work they use agents to perform. The lead author filters proposals using three admission criteria: the requested outcome must be (i) specific enough to verify with explicit checks, (ii) observable in a clean project without inspecting a solution, and (iii) implementable through a declared deliverable. When a gameplay behavior is suitable for both C++ and Blueprint, it can become a matched pair with different artifact requirements.

The team refines each accepted proposal into a task package that includes instructions, starting project, grading criteria, and other necessary information for agent to start the implementation. We verify that there are at least one reference solution that passes all required checks and that the unmodified starting project fails. Appendix~\ref{app:worked} follows one task from its instruction to the final grading pass.

\section{Experimental Design}
\label{sec:setup}

\subsection{Models and tool configurations}
We evaluate Claude Sonnet~5, DeepSeek V4 Pro-0813, Gemini 3.7 Flash, GPT-5.6 Luna, GPT-5.6 Sol, Grok~4.6, and GLM-5.3 Flash. The two editor-tool configurations use \armmcpfullname{} (\armmcp{}) and \armaurafullname{} (\armaura{}). Each model attempts every task under both configurations; C++ also has a file/shell baseline without editor MCP. All configurations use Claude Code with \ue{} on Windows. Appendix~\ref{app:configuration} records model details.

\armmcp{} exposes catalog discovery and a gateway for editor operations. \armaura{} exposes operations directly, including delegated Blueprint and Python tools. The two MCPs provide different editor tools and file-editing permissions, as summarized in Table~\ref{tab:arms}. On BP and Python, their completion differences are therefore not attributable to editor tools alone (Appendix~\ref{app:configuration}).

\begin{table}[!ht]
\centering\small
\begin{tabularx}{\linewidth}{@{}L{.19\linewidth}YY@{}}
\toprule
Configuration & C++ tasks & BP and Python tasks \\
\midrule
File/shell baseline & Read, write, edit, shell; no editor MCP & Not evaluated \\
\armmcp{} & File/shell tools and editor MCP & File/shell tools and editor MCP \\
\armaura{} & File/shell tools and editor MCP & Read/search and editor MCP; no native write/edit/shell \\
\bottomrule
\end{tabularx}
\caption{Agent tool permissions. Both editor configurations can perform several kinds of project edits; tool availability is part of the evaluated configuration.}
\label{tab:arms}
\end{table}

\subsection{Task run and completion}
Each task--model--MCP combination is evaluated with one scored attempt. The agent starts in a fresh project with the task prompt and is prompted to conduct the task. The prompt instructs agents to perform at most one final PIE playtest, which the harness does not enforce. We added this instruction after observing repeated playtest loops in preliminary runs. Agents receive no human feedback during authoring.

We report pass@1 with a 40-minute authoring budget. An attempt is successful if the agent finishes within this budget and its final submission passes all checks. Authoring time is measured from an agent launch to its exit, and the project setup and grading time are not included. Sessions normally have a 60-minute execution ceiling; final submissions returned after the 40-minute budget but under the ceiling are retained for analysis.

\subsection{Paired comparisons and observed workflows}
For each model, we compare its completion count under the two MCP settings and, on C++, also against the file/shell baseline. We also identify tasks that succeed in one setting and fail in the other, since equal totals can contain successes in different tasks. The ten paired gameplay specifications compare C++ and BP implementations of the same requested behavior. We analyze the interaction logs alongside the final test reports to examine how agents conduct their work. We also measure authoring time and the efficiency of tool calls.

We compare completion using only the gameplay tests to see whether Blueprint’s extra asset checks account for its lower score. Appendix~\ref{app:analysis-audit} provides the analysis details.

\section{Results}

\label{sec:results}

We report task completion for each model and configuration, then examine the saved submissions and tool requests to understand incomplete gameplay and the work performed during authoring. Completion uses the \scoredminutes{}-minute budget; trajectory and runtime diagnostics examine the full recorded sessions and final submissions. Appendix~\ref{app:cpp-resources} reports authoring time and tool calls.

Appendix Figure \ref{fig:completion-models} shows the full task result. C++ completion is 180/231 (77.9\%) with file/shell tools, 199/231 (86.1\%) with \armmcp{}, and 203/231 (87.9\%) with \armaura{}. Both MCP configurations complete 110/175 BP runs (62.9\%). Python completion is 62/84 (73.8\%) and 66/84 (78.6\%).

\subsection{C++ and Blueprint implementations of the same gameplay}
\label{sec:results-main}

\begin{table}[!ht]
\centering\small
\begin{tabular}{@{}lrr@{}}
\toprule
Configuration & C++ completion & BP completion \\
\midrule
\armcontrol{} & 45/70 (64.3\%) & --- \\
\armmcp{} & 46/70 (65.7\%) & 25/70 (35.7\%) \\
\armaura{} & 50/70 (71.4\%) & 20/70 (28.6\%) \\
\bottomrule
\end{tabular}
\caption{Completion within \scoredminutes{} minutes on the same ten gameplay tasks. Seven models run each task, giving 70 runs per evaluated configuration and variant.}
\label{tab:trajectory-matched}
\end{table}

\begin{figure}[t]
\centering
\includegraphics[width=\linewidth]{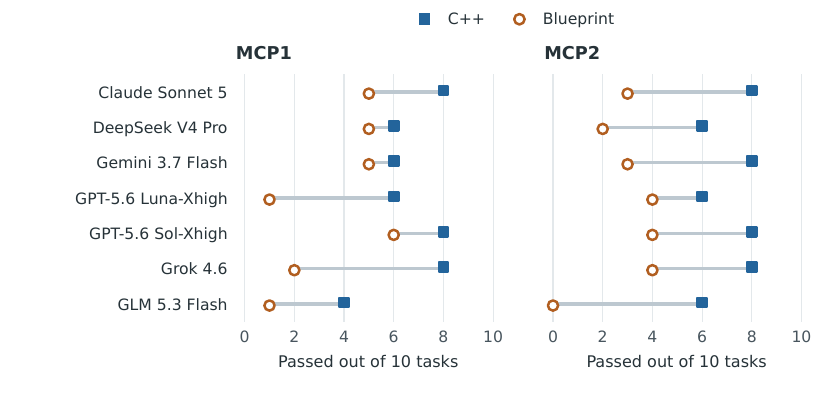}
\caption{C++ and BP completions within \scoredminutes{} minutes on the ten paired gameplay tasks. Each line connects the same model's totals under one MCP configuration, with one attempt per task variant.}
\label{fig:revision-matched}
\end{figure}

The ten paired gameplay tasks compare the same requested behavior across C++ and BP (Table~\ref{tab:trajectory-matched}). \armmcp{} completes 46/70 C++ runs (65.7\%) and 25/70 BP runs (35.7\%); \armaura{} completes 50/70 (71.4\%) and 20/70 (28.6\%), respectively. Every model completes more C++ versions under both configurations (Figure~\ref{fig:revision-matched}; Table~\ref{tab:trajectory-matched-models}). The C++ file/shell baseline already completes 45/70 runs (64.3\%), so the higher C++ completion also occurs without editor MCP on the C++ side; most of each MCP configuration's advantage over that baseline is on C++ tasks without BP counterparts (Table~\ref{tab:cpp-decomposition}).

The C++ completion advantage is 30.0 percentage points under \armmcp{} and 42.9 under \armaura{}. Counting submissions that pass the runtime checks without requiring the additional BP asset checks leaves all four completion counts unchanged (Table~\ref{tab:runtime-criterion}). Within the deadline, no BP submissions pass the runtime checks but fail the asset checks. The BP deficit persists when completion is assessed by gameplay testing. Producing saved native assets remains part of the BP authoring task.

The submissions show how the successful implementations were authored. Although some C++ prompts also permit Blueprint assets, every successful matched C++ submission under either MCP configuration changes C++ source files without changing Blueprint assets. Successful BP submissions change assets without changing the source.

The BP breakdown separates local edits from complete gameplay features (Appendix Figure~\ref{fig:bp-task-groups}). The three basic graph tasks ask for logging, arithmetic, or delayed movement; two local repair tasks target specified defects in existing behaviors. Both configurations complete 34/35 runs (97.1\%) on these five tasks. On ten asset-configuration tasks, completion is 51/70 (72.9\%) and 56/70 (80.0\%); on the ten gameplay tasks, it falls to 25/70 (35.7\%) and 20/70 (28.6\%). Agents often complete the included local operations, while fewer submissions implement all parts of a gameplay feature correctly.

\subsection{Where Blueprint implementations fall short}

On the ten paired gameplay tasks, seven additional \armmcp{} BP runs and ten additional \armaura{} runs yield passing submissions after the 40-minute deadline. Across all final submissions, those that pass asset checks but fail an explicit runtime assertion account for 24/70 (34.3\%) under \armmcp{} and 28/70 (40.0\%) under \armaura{}. Restricted to submissions whose authoring finished within \scoredminutes{} minutes, they are \revisionvalue{UnrealRuntimeFail}/\revisionvalue{UnrealAssetEligible} and \revisionvalue{McpBRuntimeFail}/\revisionvalue{McpBAssetEligible} of the on-time submissions that pass asset checks. Appendix~\ref{app:runtime-failures} lists the failed checks and the available reports.

The runtime reports provide further observations about these incomplete features. Among submissions that pass asset checks but fail runtime checks, 13 submissions across four BP tasks successfully activate an ability but fail to produce its required effect or meet the required resource check. Seven of them concern the gliding task: the character must descend more slowly while consuming Power. All seven abilities activate, and six consume Power, but none produce the required glide. For example, in three implementations, Power always activates at the start but does not exhaust after gliding for a set time. The other cases comprise four missing periodic burns, one second jump without its Power cost, and one damage operation that leaves Health unchanged. Nine of these runs finish authoring within the scoring budget. Appendix~\ref{app:runtime-failures} identifies the 13 submissions and the runtime observations supporting this breakdown.

\subsection{How Agents Inspect and Verify Their Work}

\label{sec:model-profiles}

We examine how agents respond to editing feedback and check their work under the instruction to perform at most one PIE playtest. \emph{Asset readback} means retrieving an asset's current properties or graph from the editor. \emph{PIE state readback} means retrieving values from a running game, such as character position or remaining Power. Both are actions during authoring; the harness applies the task's runtime checks to the submitted project afterward.

\textbf{Agents revise both rejected and accepted edits.} Of the 175 \armmcp{} BP runs, 95 directly write Blueprint graphs; 75 read task assets before their first graph write. Among the 58 runs with graph-writing errors, 37 responded in their next tool call by changing the request for the same graph. Agents also revised graphs after requests that returned without an error: this occurred in 30 of the 94 runs with such responses. Before making these later changes, agents often inspected the graph or requested compilation. Appendix~\ref{app:process-synthesis} gives the detailed counts and sequence definitions.

\textbf{Asset readbacks are more frequent than PIE state readbacks.} Most BP runs contain asset readback or compilation requests, while PIE state readback appears in 24/175 \armmcp{} and 30/175 \armaura{} runs (Table~\ref{tab:process-checks-bp}). Agents obtain these values through property queries, scripts that combine actions and measurements, and game-state recordings. C++ agents also use shell commands to request game or engine-test execution, in 57/231 \armmcp{} and 39/231 \armaura{} runs, including unsuccessful attempts. In ten Python \armaura{} runs across six tasks and five models, agents find the relevant API or tool, run an editor script that reports success, and read back task-asset state. The analysis follows these different routes through the request arguments and responses (Appendix~\ref{app:process-routes}).

\begin{table}[!htbp]
\centering\small
\begin{tabular}{@{}lrr@{}}
\toprule
Observed method in BP runs & \armmcp{} & \armaura{} \\
\midrule
Task-asset readback & 149/175 & 165/175 \\
Compilation request & 131/175 & 132/175 \\
PIE start request & 38/175 & 44/175 \\
Image-capture request & 15/175 & 10/175 \\
PIE state readback & 24/175 & 30/175 \\
\bottomrule
\end{tabular}
\caption{Checking actions in BP trajectories, counted over full authoring sessions. Rows overlap. Request counts include unsuccessful attempts; readback counts require returned values. PIE starts include tool calls and starts within scripts. Image requests can capture either the editor or gameplay. Appendix~\ref{app:process-routes} defines the categories and their coverage.}
\label{tab:process-checks-bp}
\end{table}

\textbf{Success revisions do not guarantee correct gameplay.} In 39 MCP1 Blueprint runs that finished within 40 minutes, every graph-writing error was followed by a changed request for the same graph that returned without an error. Nevertheless, 16 submissions failed the final evaluation, including 15 that failed runtime assertions across six models and eight tasks. A separate inspection of final submissions found five MCP1 Blueprint submissions with resource references that the engine could not resolve; these submissions also failed runtime checks (Appendix~\ref{app:attribute-binding}).

On the ten paired BP tasks, five \armmcp{} and ten \armaura{} runs contain PIE state readback and pass the asset checks, yet fail the final runtime checks. Reading a value during PIE can leave other required behaviors untested.

\section{Discussion}

\label{sec:discussion}

\paragraph{Design tasks with interacting gameplay requirements.}

A glide feature must slow descent, consume Power, and restore normal falling when Power is exhausted. These requirements need to hold together: checking resource consumption alone would not detect an implementation that leaves descent unchanged. The runtime reports and graph-revision results show why individual editing operations are not sufficient evidence of completion. For bounded gameplay tasks, evaluation should cover each required behavior and the transitions between them within one gameplay mechanism.

\paragraph{Implications for editor-native assets.}

Our findings motivate evaluating native assets in the engine systems that use them. For animation assets, materials, and scene configuration, the question is not only whether their saved properties are correct, but whether they produce the requested motion, appearance, or interaction. CraftBench-UE examines this connection in Unreal gameplay. Whether the same failure patterns occur in other engines and asset workflows remains a question for future work.

\paragraph{Verification inside the game engine.}

The observed workflows suggest two complementary forms of feedback. Inspecting asset properties and compiling graphs helps agents check what they have built; running the game and measuring its state helps them check what it does. Asset readbacks are more frequent than PIE state readbacks in the retained trajectories, measured under the prompt's instruction to perform at most one PIE playtest (\S\ref{sec:setup}). The runtime reports show several errors that can be identified using this method, such as unresolved references and missing or incorrect resource usage. Because game development offers many verification methods, such as screenshots, console logs, and PIE tests, how agents choose or combine these methods is worth studying with a larger sample size, especially when many methods are hints or stimuli rather than direct test feedback. For example, console logs in major game engines can be helpful but may also include unrelated information that bloats the agent context. For agent design, this suggests connecting each requested expected behavior to a set of verification methods and expected results, with a verification flow that connects these individual parts and automatically play-tests the implementations.

\section{Conclusion}
\label{sec:conclusion}

\benchname{} evaluates coding agents through their saved source and assets and the behavior those deliverables produce in Unreal Engine. Its harness reconstructs submissions in fresh projects and applies task-specific build, asset, and runtime checks. The \numtasks{}-task benchmark supports comparisons across three deliverable groups. On \nummatchedanalysis{} matched gameplay specifications, C++ completion exceeds BP completion under both editor configurations. Submissions that pass asset checks can still fail runtime assertions, demonstrating the value of evaluating structural requirements and gameplay behavior together.

\section{Limitations}
\label{sec:limitations}

\textbf{Task and check scope.} The \numtasks{} tasks are selected for explicit, checkable outcomes rather than sampled representatively from game development. They do not measure visual appeal or the integration of many features into a complete game. Future work could extend the task set to long-horizon projects and creative requirements with separately validated evaluation criteria. Meanwhile, our current checks include three necessary types of tests in Unreal Engine, while future work could explore custom tests to verify generated features, potentially providing more coverage than this harness.

\textbf{Experimental scope.} Each task--model--configuration run has one attempt, with at most one PIE playtest during authoring. Results do not estimate repeated-attempt variability, long-horizon task performance, or performance with unrestricted testing. Future work could explore more about how agents verify themselves to produce the expected outcome.

\textbf{Environment coverage.} All experiments use \ue{} on Windows through Claude Code. We have not tested whether the observed completion patterns persist across other engine versions, operating systems, or agent runtimes.

% ICLR 2027 exempts ethics, reproducibility, and AI-use statements from
% the main-text limit. Start these on a fresh page in the submission.
\ifanon\clearpage\fi
% Team acknowledgments appear only in the named paper.
\ifanon\else
\section*{Acknowledgments}
We thank the entire \productname{} team for their support and contributions to the benchmark tasks.
\fi

\section*{Ethics Statement}
\label{sec:ethics}
The authors' organization develops \armaura{}, whose implementation remains proprietary. Appendix~\ref{app:configuration} describes its exposed tool surface. The benchmark team authors the tasks and verifiers. Both editor configurations are evaluated on the same task suite using the same verifiers, and their results are reported separately.

\section*{Reproducibility Statement}
\label{sec:repro}
The harness, benchmark materials, and analysis code will be released upon publication\ifanon.\else{} at \artifacturl.\fi{} Task instructions, starting-project configurations, verifier, and reference implementations are included as well. Appendix~\ref{app:trajectory-method} defines the trajectory analyses. Appendix~\ref{app:release} lists the release contents. Users will need to install Unreal Engine and provide it in the environment; engine binaries and proprietary MCP internals are not redistributed.

\section*{AI Use Statement}
\label{sec:ai-disclosure}

Generative-AI assistants supported task ideation, methodological critique, data analysis, and figure brainstorming. The rule-based pipeline in \S\ref{sec:benchmark} produced benchmark verdicts. The authors remain responsible for the analysis, claims, and final manuscript.

\bibliographystyle{iclr2027_conference}
\bibliography{references}

\appendix
\section{A Worked Example of Task Validation}
\label{app:worked}

\begin{figure}[!ht]
\centering
\includegraphics[width=\linewidth]{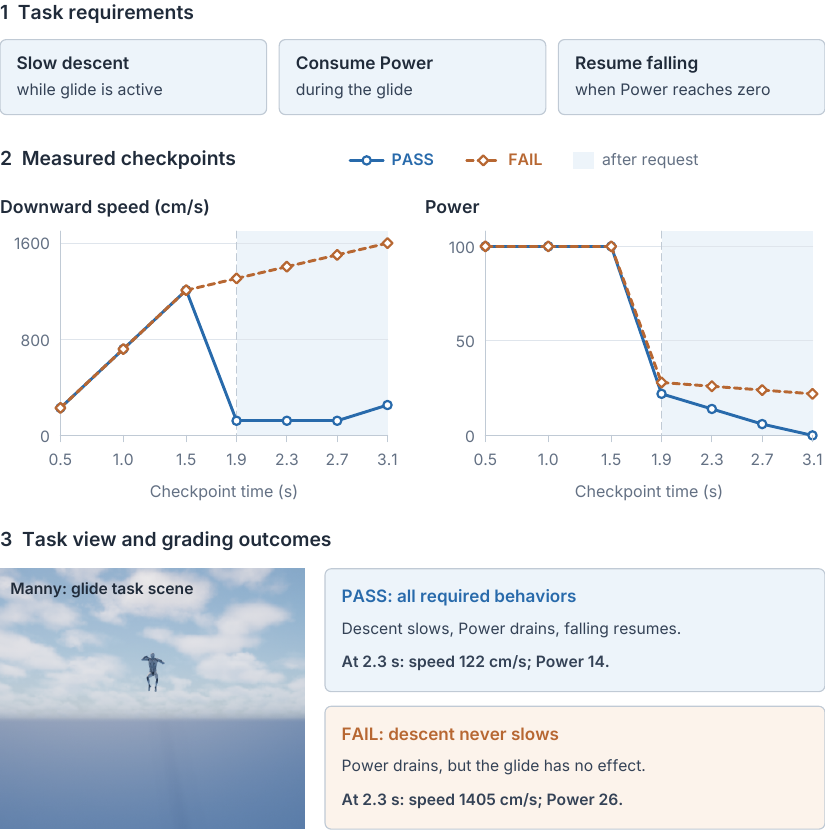}
\caption{Archived development/validation traces for the glide task, separate from the study's agent runs. In the shaded post-request interval, the passing trace slows descent while Power drains; the failing trace drains Power without slowing descent. The plotted values and verdict labels are transcribed from a retained development capture. The Manny screenshot is a separate scene illustration, not a frame from either trace.}
\label{fig:glide-example}
\end{figure}

The glide task in Figure~\ref{fig:glide-example} asks the agent to slow a falling character while consuming Power, then restore normal falling when Power runs out. The runtime fixture requests ability activation and observes descent and Power at scheduled checkpoints. Checking Power alone would miss an implementation that consumes the resource without slowing the character.

This example illustrates why task authors check behavior after integration into the engine. Figure~\ref{fig:task-overview} in the main paper illustrates tasks across the three deliverable groups.

\section{Execution Configuration and Submission Handling}
\label{app:harness}
\label{app:configuration}

The experiments use \ue{} on Windows with Claude Code as the agent driver. Table~\ref{tab:analysis-model-ids} records the requested outer-agent model identifiers. Provider-qualified identifiers are served via the same Claude Code CLI, with its Anthropic base URL pointing to OpenRouter's Anthropic-compatible endpoint and the OpenRouter key supplied as the CLI's bearer token; \texttt{claude-sonnet-5} is requested directly from Anthropic. The analysis manifest retains available model and driver metadata.

\begin{table}[!ht]
\centering\small
\begin{tabular}{@{}ll@{}}
\toprule
Display label & Requested model identifier \\
\midrule
Claude Sonnet 5 & \texttt{claude-sonnet-5} \\
DeepSeek V4 Pro & \texttt{deepseek/deepseek-v4-pro-0813} \\
Gemini 3.7 Flash & \texttt{google/gemini-3.7-flash} \\
GPT-5.6 Luna & \texttt{openai/gpt-5.6-luna} \\
GPT-5.6 Sol & \texttt{openai/gpt-5.6-sol} \\
Grok 4.6 & \texttt{x-ai/grok-4.6} \\
GLM-5.3 Flash & \texttt{z-ai/glm-5.3-flash} \\
\bottomrule
\end{tabular}
\caption{Requested agent model identifiers.}
\label{tab:analysis-model-ids}
\end{table}

\paragraph{Tool exposure.}
\armmcp{} exposes three entry points: catalog discovery, tool description, and a gateway that calls a named operation. \armaura{} exposes 176 entry points, including direct editing and delegated tools. These names include asset and Blueprint inspection, Blueprint graph editing and compilation, asset creation and editing, text and C++ file editing with in-editor Python execution, play-in-editor control, and more. A gateway can access many operations, so entry-point counts do not measure tool coverage. Table~\ref{tab:arms} gives file and shell permissions. Initialization failures count as failed attempts. 

\paragraph{Timing and comparison scope.}
The runner used a 60-minute execution ceiling. Completion required the agent to finish within 40 minutes and its final submission to pass all required checks. The submitted project is graded afterward. The C++ baseline uses two hosts, whereas its MCP comparisons use one. Both hosts use the same hardware configuration, each with an NVIDIA GeForce RTX 5080 GPU. On BP and Python tasks, \armmcp{} also retains the native write, edit, and shell tools that \armaura{} does not (Table~\ref{tab:arms}), and some \armmcp{} runs on these surfaces issue native file or shell writes, so this permission difference is exercised in the recorded sessions. Driver versions vary within and across task groups, and some records lack version metadata. These differences limit attribution of completion or timing differences to MCP access alone.

Behavior (L2) checks launch the editor with \texttt{-deterministic -FPS=60}, which fixes the simulation step at 1/60 second from the first frame; a task may declare further legs at other fixed steps, which are replayed in separate sessions and must all pass. Checkpoints are evaluated against the world's game time, counted from when play begins rather than from when the test fixture starts.

\paragraph{Agent inputs and testing.}

Each run supplies the task prompt and starting project, specifies where to save changes, and instructs the agent to perform at most one PIE playtest. The limit is prompt-level only: the harness does not intercept or refuse play-start requests during authoring, so additional playtests consume the run's authoring budget rather than being blocked; however, there is no such request in the current trajectories. Play-start requests are instead recovered from the trajectories after the run (Appendix~\ref{app:process-routes}). The agent receives no human feedback or grading reports during authoring. The harness grades its saved submission afterward.

\paragraph{Saved submissions.}
The harness collects changed or new files accepted by the starting project's submission manifest and applies them to a fresh starting project. The grader applies the task's additional file and configuration constraints before evaluating the reconstructed project. This procedure makes the saved submission, rather than unsaved editor state, the object of evaluation.

\subsection{Access to Evaluation Material}
\label{app:hiding}

Before authoring, the harness withholds reference implementations, verifier sources, and repository metadata that could expose them. It also removes other tasks' scaffolds and replaces runtime test bodies with stubs while retaining the build definitions needed to compile the authoring project. Withheld material is restored for grading in the clean project.

The harness monitors the withheld paths and excludes an attempt from scoring if it detects exposed evaluation material. Task instructions and permitted starting content remain available during authoring.

\subsection{Release Materials}
\label{app:release}

The release includes the harness, task instructions, starting-project setup, verifiers, reference implementations, retained submissions, and analysis code. Run identifiers and source hashes connect the reported panel, check results, and trajectory classifications to their underlying records. Public trajectories omit credentials and private content while retaining the feedback and submission evidence needed to inspect the analyses. Users provide an Unreal Engine installation; engine binaries and proprietary \armaura{} internals are not redistributed.

\section{Detailed Results and Scoring Checks}
\label{app:analysis-audit}
\label{app:verdicts}

\begin{figure}[!htbp]
\centering
\includegraphics[width=\linewidth]{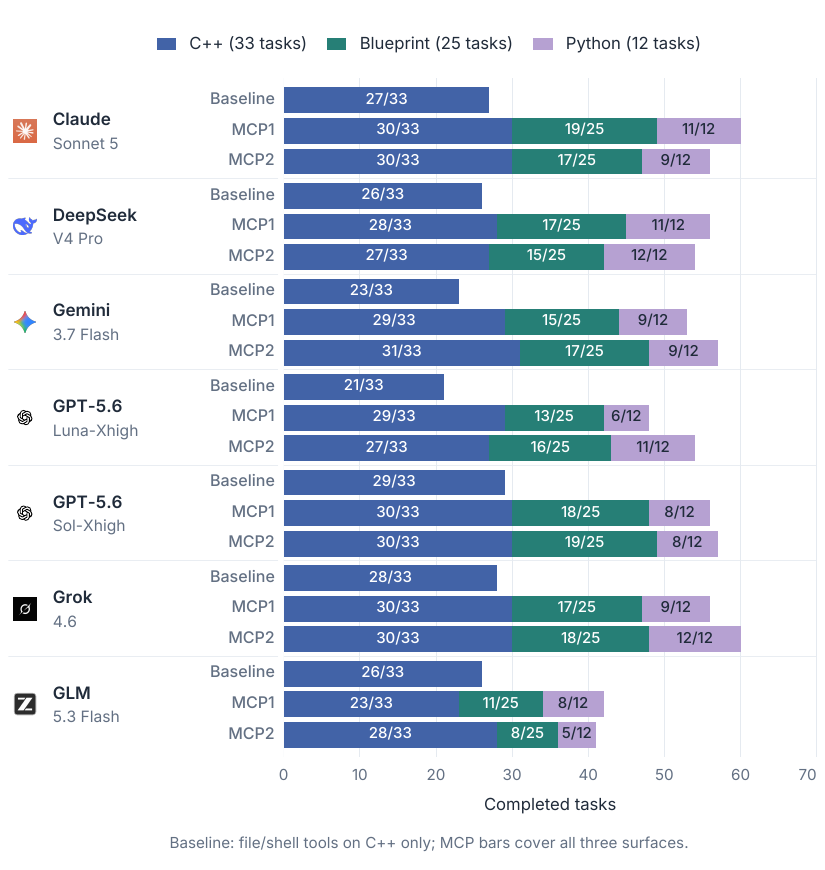}
\caption{Task completion within \scoredminutes{} minutes, by model and configuration. Each model has three stacked bars: Baseline, \armmcp{}, and \armaura{}. Colored segment lengths indicate the counts of completed C++, Blueprint, and Python tasks. Labels inside each segment give completed tasks out of 33 C++, 25 Blueprint, or 12 Python tasks. The baseline uses file/shell tools and was evaluated only on C++, so its total bar length is not comparable with those of the MCP configurations. Table~\ref{tab:analysis-models-40m} gives the same counts.}
\label{fig:completion-models}
\end{figure}

This appendix gives detailed results for the 70-task suite shown in Figure~\ref{fig:completion-models} and its ten paired gameplay specifications. Each record is identified by task, model, and configuration. The resulting panel has no missing or duplicate combinations. Completion requires \texttt{overall=PASS} and meets the 40-minute time budget. Build checks are L1, runtime checks of gameplay are L2, and asset checks of saved structure are L2I.

\begin{table}[!ht]
\centering\footnotesize
\begin{tabular}{@{}lrrr@{}}
\toprule
\multicolumn{4}{c}{\textbf{C++}} \\
Model & \armcontrol{} & \armmcp{} & \armaura{} \\
\midrule
Claude Sonnet 5 & 27/33 (81.8\%) & 30/33 (90.9\%) & 30/33 (90.9\%) \\
DeepSeek V4 Pro & 26/33 (78.8\%) & 28/33 (84.8\%) & 27/33 (81.8\%) \\
Gemini 3.7 Flash & 23/33 (69.7\%) & 29/33 (87.9\%) & 31/33 (93.9\%) \\
GPT-5.6 Luna & 21/33 (63.6\%) & 29/33 (87.9\%) & 27/33 (81.8\%) \\
GPT-5.6 Sol & 29/33 (87.9\%) & 30/33 (90.9\%) & 30/33 (90.9\%) \\
Grok 4.6 & 28/33 (84.8\%) & 30/33 (90.9\%) & 30/33 (90.9\%) \\
GLM 5.3 Flash & 26/33 (78.8\%) & 23/33 (69.7\%) & 28/33 (84.8\%) \\
\bottomrule
\end{tabular}
\par\medskip
\begin{tabular}{@{}lrr@{}}
\toprule
\multicolumn{3}{c}{\textbf{BP}} \\
Model & \armmcp{} & \armaura{} \\
\midrule
Claude Sonnet 5 & 19/25 (76.0\%) & 17/25 (68.0\%) \\
DeepSeek V4 Pro & 17/25 (68.0\%) & 15/25 (60.0\%) \\
Gemini 3.7 Flash & 15/25 (60.0\%) & 17/25 (68.0\%) \\
GPT-5.6 Luna & 13/25 (52.0\%) & 16/25 (64.0\%) \\
GPT-5.6 Sol & 18/25 (72.0\%) & 19/25 (76.0\%) \\
Grok 4.6 & 17/25 (68.0\%) & 18/25 (72.0\%) \\
GLM 5.3 Flash & 11/25 (44.0\%) & 8/25 (32.0\%) \\
\bottomrule
\end{tabular}
\par\medskip
\begin{tabular}{@{}lrr@{}}
\toprule
\multicolumn{3}{c}{\textbf{Python}} \\
Model & \armmcp{} & \armaura{} \\
\midrule
Claude Sonnet 5 & 11/12 (91.7\%) & 9/12 (75.0\%) \\
DeepSeek V4 Pro & 11/12 (91.7\%) & 12/12 (100.0\%) \\
Gemini 3.7 Flash & 9/12 (75.0\%) & 9/12 (75.0\%) \\
GPT-5.6 Luna & 6/12 (50.0\%) & 11/12 (91.7\%) \\
GPT-5.6 Sol & 8/12 (66.7\%) & 8/12 (66.7\%) \\
Grok 4.6 & 9/12 (75.0\%) & 12/12 (100.0\%) \\
GLM 5.3 Flash & 8/12 (66.7\%) & 5/12 (41.7\%) \\
\bottomrule
\end{tabular}
\par\medskip
\caption{Full-task completions within \scoredminutes{} minutes, with counts and percentages shown separately for each surface. The file/shell baseline is evaluated only on C++.}
\label{tab:analysis-models-40m}
\end{table}

\begin{table}[!ht]
\centering\small
\begin{tabular}{@{}lrrrr@{}}
\toprule
Model & \multicolumn{2}{c}{\armmcp{}} & \multicolumn{2}{c}{\armaura{}} \\ & C++ & BP & C++ & BP \\
\midrule
Claude Sonnet 5 & 8/10 & 5/10 & 8/10 & 3/10 \\
DeepSeek V4 Pro & 6/10 & 5/10 & 6/10 & 2/10 \\
Gemini 3.7 Flash & 6/10 & 5/10 & 8/10 & 3/10 \\
GPT-5.6 Luna & 6/10 & 1/10 & 6/10 & 4/10 \\
GPT-5.6 Sol & 8/10 & 6/10 & 8/10 & 4/10 \\
Grok 4.6 & 8/10 & 2/10 & 8/10 & 4/10 \\
GLM 5.3 Flash & 4/10 & 1/10 & 6/10 & 0/10 \\
\bottomrule
\end{tabular}
\caption{Per-model completion within \scoredminutes{} minutes on the ten paired gameplay specifications.}
\label{tab:trajectory-matched-models}
\end{table}

% Generated by analysis/iteration_20260909/build_tables.py
\begin{table}[!ht]
\centering\small
\begin{tabular}{@{}lrrr@{}}
\toprule
C++ task subset & Baseline & \armmcp{} & \armaura{} \\
\midrule
Ten paired gameplay tasks & 45/70 & 46/70 & 50/70 \\
Other 23 C++ tasks & 135/161 & 153/161 & 153/161 \\
\bottomrule
\end{tabular}
\caption{C++ completion within \scoredminutes{} minutes, split by whether the specification also has a BP version. The split follows the task definitions. It is not selected from outcomes.}
\label{tab:cpp-decomposition}
\end{table}

\paragraph{BP task groups.}
We group BP tasks by their instructions: three basic graph tasks (logging, arithmetic, and delayed movement), two local repairs (door hitch and inventory stacking), ten asset-configuration tasks, and ten paired gameplay tasks. Asset-configuration tasks check saved structure; gameplay tasks also check runtime behavior.

\begin{figure}[!ht]
\centering
\includegraphics[width=\linewidth]{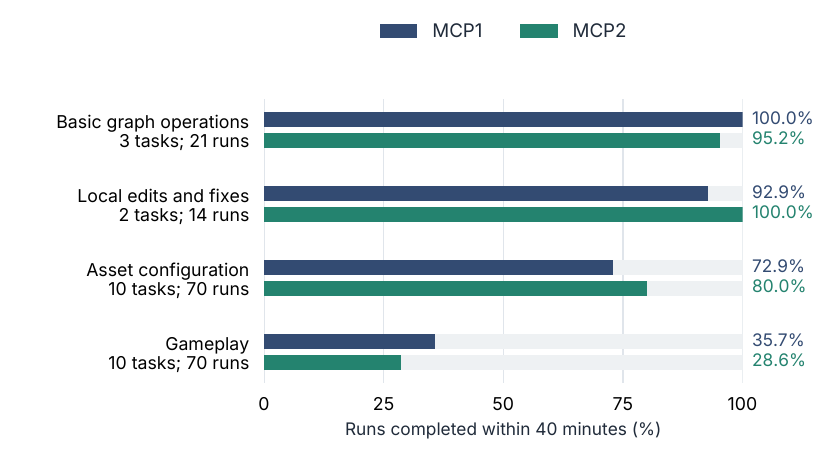}
\caption{BP completion by task type. Row labels give the number of tasks and runs per configuration (seven models per task). Groups follow the task instructions and differ in content, starting state, and required checks.}
\label{fig:bp-task-groups}
\end{figure}

\paragraph{Counting runtime passes.}
The ten C++/BP gameplay pairs share runtime fixtures; BP also requires asset checks. Counting recorded runtime passes within 40 minutes leaves all four completion counts unchanged (Table~\ref{tab:runtime-criterion}). Every counted runtime pass also passes the build checks.

\begin{table}[!ht]
\centering\small
\begin{tabular}{@{}lrrrr@{}}
\toprule
Configuration & C++ full & BP full & BP runtime pass & Gap (pp) \\
\midrule
\armmcp{} & 46/70 & 25/70 & 25/70 & 30.0 \\
\armaura{} & 50/70 & 20/70 & 20/70 & 42.9 \\
\bottomrule
\end{tabular}
\caption{Completion on the ten paired gameplay tasks, with and without the BP asset checks. Each task group has 70 attempts per configuration. All counts require authoring to finish within 40 minutes. Counting recorded BP runtime passes leaves the full-task counts and the C++--BP gap unchanged; the original submissions are used in both counts. Gap is measured in percentage points.}
\label{tab:runtime-criterion}
\end{table}

All 20 BP submissions that fail asset checks have runtime reports, including five \armmcp{} and seven \armaura{} submissions returned within 40 minutes. Five attempts have \texttt{FAIL\_NO\_EDITS} and no runtime outcome; they remain failures in the denominator. One \armmcp{} submission passes runtime but fails asset checks after 52.1 minutes. Runtime automation failures remain failures even when individual messages report successful behavior.

\paragraph{Python input bindings.}
The input-binding task checks action types, six key bindings, and their modifiers across four input assets. Scores use the engine's active \texttt{DefaultKeyMappings.Mappings} array in the retained submissions. These checks assess saved input configuration without simulating keypresses.

\section{Trajectory Analysis}
\label{app:trajectory-method}

\subsection{Inspection and Verification in Agent Trajectories}
\label{app:behavior-patterns}

The trajectory analysis covers 693 C++ transcripts, 350 BP transcripts, and 168 Python transcripts, using the run membership and reported outcomes in Appendix~\ref{app:analysis-audit}. Counts cover full authoring sessions unless a time limit is stated. For \armmcp{}, the operation is read from the gateway request's toolset and operation arguments. Requests are matched to returns by tool-use identifier and occurrence; seven requests lack retained returns, and six nested calls in the baseline C++ traces are excluded. The counts describe visible agent requests and returns; unobserved work inside tool services is outside this analysis.

\paragraph{Asset and PIE state readback.}
\label{app:self-verification}
\label{app:process-routes}
An \emph{asset readback} is a returned value from a task-owned asset, excluding errors, search results, schema descriptions, and object handles without state. A \emph{PIE state readback} contains a value measured while the game runs in Play-in-Editor (PIE). It may come from a dedicated getter, a property or transform query targeting a \texttt{UEDPIE} object, an inspected script, a combined action sequence, or a recording. A \emph{PIE start request} counts the request to start the game, including starts within identified scripts. A start request alone is insufficient for a state readback.

Table~\ref{tab:process-checks-bp} reports these methods for all BP runs; Table~\ref{tab:process-models-bp} separates them by model. Images count separately and require an image block in the agent's conversation: seven and eight runs qualify, respectively. Script-based readbacks require returned measurements rather than a printed success statement. Compilation counts requests; a return without an error, including a null return, does not establish successful compilation.

\begin{table}[!htbp]
\centering\small

\begin{tabular}{lrrrrrr}
\toprule
& \multicolumn{2}{c}{PIE start request} & \multicolumn{2}{c}{PIE state readback} & \multicolumn{2}{c}{Image return} \\ Model & \armmcp{} & \armaura{} & \armmcp{} & \armaura{} & \armmcp{} & \armaura{} \\
\midrule
Sonnet & 17 & 18 & 12 & 11 & 1 & 5 \\
DeepSeek & 4 & 6 & 2 & 4 & 2 & 0 \\
Gemini & 2 & 4 & 1 & 3 & 0 & 0 \\
Luna & 0 & 2 & 0 & 1 & 0 & 0 \\
Sol & 0 & 0 & 0 & 0 & 0 & 0 \\
Grok & 1 & 1 & 1 & 1 & 0 & 0 \\
GLM & 14 & 13 & 8 & 10 & 4 & 3 \\
\bottomrule
\end{tabular}
\caption{BP inspection methods by model, out of 25 runs per configuration. PIE start requests include identified scripts. PIE state readbacks require returned measurements from the running game. An image must appear in the agent's conversation. Rows use the definitions in Appendix~\ref{app:process-routes}.}
\label{tab:process-models-bp}
\end{table}

On C++, 151 reviewed shell requests ask for game or engine-test execution, covering 57/231 \armmcp{} and 39/231 \armaura{} runs. Two further requests only list available tests and are excluded. The counts include failed starts and background requests whose completion is unconfirmed. These shell operations are separate from PIE start requests.

On Python, ten \armaura{} runs across six tasks and five models contain tool/API discovery, a task-directed modification script, an explicit \texttt{bSuccess:true} return, and subsequent asset readback in the task's namespace. The readback need not concern the modified field. \armmcp{} has a different return protocol, so this sequence is not used to rank script success between configurations.

\paragraph{Graph revisions.}
\label{app:process-synthesis}
All 175 \armmcp{} BP trajectories are screened for direct graph writing. The 95 qualifying runs contain 454 writes with complete request--return matching; 75 read task assets before their first graph write. An \emph{accepted graph write} returns without a reported tool error. A revision changes the arguments for the same graph after the previous write returns. Six malformed writes have no recoverable graph identity and cannot be paired. \emph{Immediate} means that the revision is the next agent tool request, rather than a short elapsed time.

Table~\ref{tab:process-graph-sequences} distinguishes revisions after accepted and rejected writes, together with intervening readbacks and compilation returns. Revisions can extend functionality as well as correct errors.

\begin{table}[!htbp]
\centering\small

\begin{tabular}{@{}lrrr@{}}
\toprule
Observed sequence & All (95) & Same tasks (49) & $\leq40$ min (79) \\
\midrule
Accepted write, then changed same-graph write & 30 & 16 & 20 \\
Readback between those writes & 27 & 15 & 17 \\
Compilation return between those writes & 21 & 11 & 12 \\
Both readback and compilation return & 18 & 10 & 9 \\
Error, then immediate same-graph change & 37 & 20 & 26 \\
Acceptance, then immediate same-graph change & 4 & 1 & 4 \\
\bottomrule
\end{tabular}
\caption{Graph revisions in \armmcp{} BP runs. Columns cover all direct graph-writing runs, the seven tasks on which every model directly writes graphs, and sessions ending within 40 minutes. Rows overlap. A compilation return need not report successful compilation. Immediate means the next agent tool request after the preceding write returns; 58 runs contain errors, and 30 revise accepted writes.}
\label{tab:process-graph-sequences}
\end{table}

Table~\ref{tab:behavior-recovery} links graph-writing errors and subsequent accepted revisions to submission outcomes. Of the 16 on-time failures after every reported error receives an accepted revision, 15 fail explicit runtime checks, spanning six models and eight gameplay tasks.

\begin{table}[!htbp]
\centering
\small
\caption{Graph-writing errors and subsequent revisions in \armmcp{} BP runs. Rows successively restrict the 175-run panel. Accepted revisions return without a reported tool error. On-time passes finish authoring within 40 minutes; final passes include later submissions.}
\label{tab:behavior-recovery}
\begin{tabular}{lrrr}
\toprule
Sequence & Runs & On-time pass & Final pass \\
\midrule
Uses direct graph writing & 95 & 55 & 62 \\
Receives graph-writing error & 58 & 27 & 34 \\
Changed same-graph request accepted & 55 & 26 & 32 \\
Every error has \revadd{a} later accepted change & 50 & 23 & 29 \\
Previous row, session ends within 40 min & 39 & 23 & 23 \\
\bottomrule
\end{tabular}
\end{table}

% \paragraph{Reproduction.}
% The derived-record pack in \path{analysis/section5_revision_20260913/} reproduces these counts with a read-only script. It includes per-run classifications, request/return locations, source hashes, and the source-analysis manifest. Re-annotating requests requires the retained trajectories. The analysis excludes ambiguous operations and does not execute recorded scripts.

\subsection{Unresolved Resource References}
\label{app:attribute-binding}

Five \armmcp{} BP submissions across five tasks and two models contain both an unresolved-resource warning and a failed runtime check. Two omit the resource's owning class; three use a class-qualified name that the engine cannot resolve. The warnings establish unresolved references, but need not explain every failed check: the healing warning concerns Health, while the failed check concerns MaxHealth initialization. The census covers retained runtime reports from 101/105 \armmcp{} and 103/105 \armaura{} runs on the \auditvalue{BpDeclaredL2} BP tasks that declare runtime checks, including 66/70 and 69/70 paired gameplay runs.

UE 5.8 source inspection identifies two supported lookup routes: an owning class plus a member name, or a valid full field path. No submission was modified to test whether resolving its reference would complete the task.

\subsection{Runtime Failures After Passing Asset Checks}
\label{app:runtime-failures}
\label{app:bp-case}

Of the 140 BP attempts on matched gameplay tasks, runtime reports are retained for 66/70 under \armmcp{} and 69/70 under \armaura{}. The five attempts without reports have \texttt{FAIL\_NO\_EDITS} outcomes. We select submissions that pass asset checks and fail an explicit runtime assertion: 24/70 and 28/70. One additional \armmcp{} automation failure has no such assertion and is excluded from this category.

\begin{table}[!ht]
\centering\small
\begin{tabular}{@{}lrr@{}}
\toprule
Failure after passing asset checks & \armmcp{} & \armaura{} \\
\midrule
Initial Health or Power & 3 & 4 \\
Required effect or resource use & 8 & 6 \\
Pose, display, or object indicator & 5 & 8 \\
Restoration or preserved behavior & 5 & 6 \\
Poison damage ratio & 3 & 4 \\
\midrule Total attempts & 24/70 & 28/70 \\
Authoring finished within 40 minutes & 19/70 & 20/70 \\
\bottomrule
\end{tabular}
\caption{BP submissions that pass asset checks but fail a runtime check. Each of the 52 selected attempts has one observed failure category; subsequent checks may not execute if a previous check failed.}
\label{tab:bp-runtime-failures}
\end{table}

% Generated by analysis/iclr_revision_20260913/recount.py; do not edit.
\begin{table}[!ht]
\centering\small
\begin{tabular}{@{}lrr@{}}
\toprule
Selected BP submissions & \armmcp{} & \armaura{} \\
\midrule
On-time submissions passing asset checks & 45 & 40 \\
Runtime checks pass & 25 & 20 \\
Explicit runtime assertion failure & 19/45 (42.2\%) & 20/40 (50.0\%) \\
Runtime automation failure & 1 & 0 \\
\bottomrule
\end{tabular}
\caption{Runtime outcomes among BP submissions on the ten gameplay specifications that pass asset checks and finish authoring within \scoredminutes{} minutes. The last three rows partition this selected subset; automation failures are separate from failed assertions.}
\label{tab:revision-asset-runtime}
\end{table}

Categories group the state measured by failed assertions, with duplicate messages removed within each attempt. Each of these 52 attempts has one observed category; later checks may not execute. The reports evaluate the final saved submissions. 

Thirteen effect or resource failures outside poison stacking also show the ability got activated successfully: seven under \armmcp{} and six under \armaura{}, across area burn, double jump, glide, and Health operations. Seven concern gliding; six consume Power before any verifier-induced exhaustion. Some reports contain no valid glide sample or no forced-exhaustion stage, so these observations do not establish a common failure mechanism across all seven submissions, but show how agents might fail when implementing features. Eight versioned verifier source objects establish how activation and resource use were measured. C++ failure selection requires passing build checks; BP selection requires passing asset checks.

\section{Authoring Time and Cost}
\label{app:cpp-resources}
\suppressfloats[t]

\begin{figure}[!ht]
\centering
\includegraphics[width=0.9\linewidth]{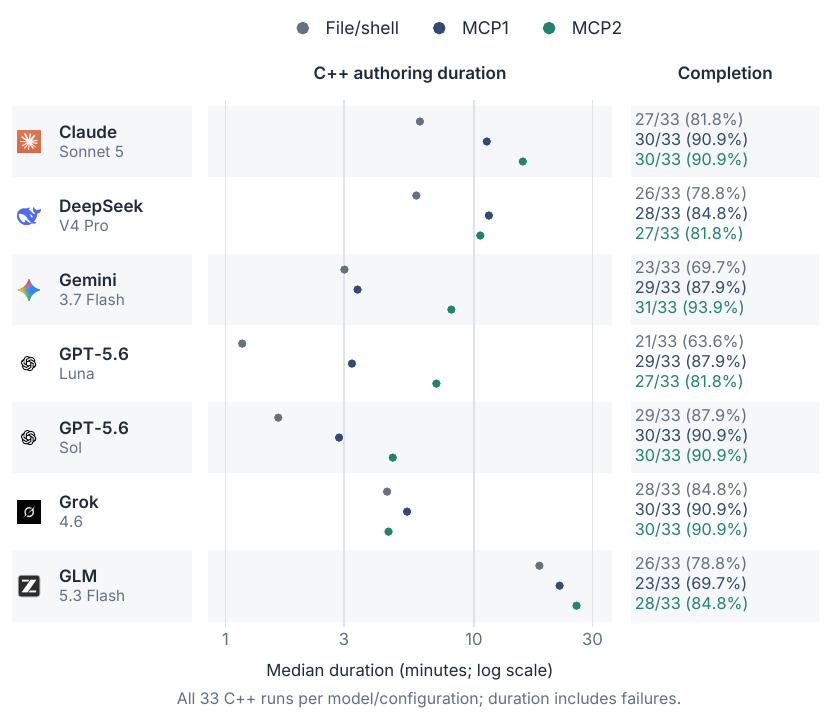}
\caption{C++ completion and median authoring time on the same 33 tasks per configuration. Markers show median duration on a logarithmic scale; labels give completion counts. All runs enter the duration calculation.}
\label{fig:cpp-authoring-time}
\end{figure}

For all seven models, median C++ authoring time is longer with either MCP configuration than with the Baseline (Figure~\ref{fig:cpp-authoring-time}). Durations include unsuccessful runs and sessions that exceed the \scoredminutes{}-minute deadline; completion still requires finishing within that deadline.

\paragraph{Input context and billed cost.}
Of 231 C++ MCP task--model pairs, 229 have identical retained task prompts and 185 have both first requests linked to billing records. In these 185 pairs, median first-request input is 3.99 times as large under \armaura{} as under \armmcp{}, a median difference of 82,354 tokens. Each configuration retains its own system instructions and tool descriptions.

Full-session billing is available for 162 prompt-matched pairs (Table~\ref{tab:cpp-inference-charges}). The median paired \armaura{}/\armmcp{} charge ratio is 0.21 for DeepSeek and 1.80--2.69 for the other five covered models. Cache use also differs by model: among 19 DeepSeek pairs with complete token records, the median share of input served from cache is 3.47\% under \armmcp{} and 98.21\% under \armaura{}; for Sol's 33 pairs, it is 92.12\% and 93.17\%.

% Generated by analysis/iteration_20260909/build_tables.py
\begin{table}[!ht]
\centering\footnotesize
\begin{tabular}{@{}lrrrrr@{}}
\toprule
Model & Pairs & \armmcp{} & \armaura{} & Charge ratio & Higher \\
\midrule
Claude Sonnet 5 & 0/33 & --- & --- & --- & --- \\
DeepSeek V4 Pro & 20/33 & 18/20 & 19/20 & 0.21$\times$ & 6/20 \\
Gemini 3.7 Flash & 32/33 & 28/32 & 31/32 & 1.86$\times$ & 23/32 \\
GPT-5.6 Luna & 29/33 & 27/29 & 25/29 & 2.48$\times$ & 29/29 \\
GPT-5.6 Sol & 33/33 & 30/33 & 30/33 & 2.69$\times$ & 33/33 \\
Grok 4.6 & 31/33 & 28/31 & 29/31 & 1.80$\times$ & 29/31 \\
GLM-5.3 Flash & 17/33 & 15/17 & 16/17 & 2.09$\times$ & 17/17 \\
\midrule C++ linked subset & 162/231 & 146/162 & 150/162 & 2.21$\times$ & 137/162 \\
\bottomrule
\end{tabular}
\caption{C++ completion and billed inference on the matched subset. Completion columns count passes within \scoredminutes{} minutes. \emph{Charge ratio} is the median paired \armaura{}/\armmcp{} ratio; \emph{Higher} counts pairs with larger \armaura{} charges. Pair counts give billing coverage out of 33 tasks per model.}
\label{tab:cpp-inference-charges}
\end{table}

\paragraph{Accounting.}
We match generation IDs from CLI summaries and assistant messages to OpenRouter billing records, require the requested model and a numeric charge for every observed generation, deduplicate IDs across four ledgers, and sum \texttt{cost\_details.upstream\_inference\_cost}. This excludes unrelated projects in the same ledgers. On fully linked C++ runs, CLI cost estimates exceed charges by model-dependent median factors of 1.25--54.85; we use the billed amounts directly, without a conversion factor. Charges cover full agent sessions, including requests after the deadline, and exclude engine compute and unobserved inference inside tools. BP, Python, and direct-Anthropic Sonnet runs have no usable billing linkage and are omitted from cost comparisons.

Input totals add the ledger's separate uncached, cache-read, and cache-write fields; cache shares use that total as the denominator. The result-summary input field already includes cached tokens and is never added to its cache fields.
\section{Task Catalog and Paired Gameplay Specifications}
\label{app:catalog}
\label{app:corpus}
\label{app:extension}
Table~\ref{tab:task-catalog} lists every task in the primary study, its starting project, and its required checks. The \numthirdperson{} Third Person tasks use Unreal's standard template; the other \numtemplateproj{} use the CraftBench template. Task names are directory identifiers so that a reader can locate the instructions and verifier directly. Table~\ref{tab:matched-task-design} gives the requested behaviors in the \nummatchedpairs{} C++/BP pairs. Their task definitions name the same interface and verifier, while BP additionally requires saved BP graphs rather than C++.

The task bank also contains \numextensiontasks{} additional tasks outside the primary study that were excluded due to the time and budget limit. They do not enter the reported completion rates or trajectory counts.

% Generated from analysis/task_catalog_20260908.json; check columns are task requirements.
\begingroup
\small
\setlength{\tabcolsep}{4pt}
\begin{longtable}{@{}L{.05\linewidth}L{.45\linewidth}L{.16\linewidth}ccc@{}}
\caption{The 70 study tasks, with their starting projects and required grading gates. IDs refer to task directories. Build, runtime, and inspect denote L1, L2, and L2I; $\checkmark$ marks a required check and --- marks one not required. The C++, BP, and Python groups contain 33, 25, and 12 tasks.}\label{tab:task-catalog}\\
\toprule
Row & Task ID & Starting project & Build & Runtime & Inspect \\
\midrule
\endfirsthead
\multicolumn{6}{l}{\tablename\ \thetable{} (continued)}\\
\toprule
Row & Task ID & Starting project & Build & Runtime & Inspect \\
\midrule
\endhead
\midrule
\multicolumn{6}{r}{Continued on next page}\\
\endfoot
\bottomrule
\endlastfoot
\multicolumn{6}{@{}l}{\textbf{C++ source tasks}}\\[2pt]
C01 & \texttt{gp-\allowbreak{}crafting-\allowbreak{}queue} & CraftBench template & $\checkmark$ & $\checkmark$ & --- \\
C02 & \texttt{gp-\allowbreak{}dot-\allowbreak{}aoe-\allowbreak{}burn-\allowbreak{}cpp} & Third Person & $\checkmark$ & $\checkmark$ & --- \\
C03 & \texttt{gp-\allowbreak{}double-\allowbreak{}jump-\allowbreak{}stamina-\allowbreak{}cpp} & Third Person & $\checkmark$ & $\checkmark$ & --- \\
C04 & \texttt{gp-\allowbreak{}glide-\allowbreak{}stamina-\allowbreak{}cpp} & Third Person & $\checkmark$ & $\checkmark$ & --- \\
C05 & \texttt{gp-\allowbreak{}harvestable-\allowbreak{}regrow} & CraftBench template & $\checkmark$ & $\checkmark$ & --- \\
C06 & \texttt{gp-\allowbreak{}heal-\allowbreak{}over-\allowbreak{}time-\allowbreak{}cpp} & Third Person & $\checkmark$ & $\checkmark$ & --- \\
C07 & \texttt{gp-\allowbreak{}health-\allowbreak{}attribute-\allowbreak{}ops-\allowbreak{}cpp} & Third Person & $\checkmark$ & $\checkmark$ & --- \\
C08 & \texttt{gp-\allowbreak{}inventory-\allowbreak{}stacking} & CraftBench template & $\checkmark$ & $\checkmark$ & --- \\
C09 & \texttt{gp-\allowbreak{}poison-\allowbreak{}dot-\allowbreak{}stack-\allowbreak{}cpp} & Third Person & $\checkmark$ & $\checkmark$ & --- \\
C10 & \texttt{gp-\allowbreak{}spawner-\allowbreak{}population} & CraftBench template & $\checkmark$ & $\checkmark$ & --- \\
C11 & \texttt{t0-\allowbreak{}sanity-\allowbreak{}log-\allowbreak{}on-\allowbreak{}beginplay} & CraftBench template & $\checkmark$ & $\checkmark$ & --- \\
C12 & \texttt{t1-\allowbreak{}data-\allowbreak{}asset-\allowbreak{}drives-\allowbreak{}speed} & CraftBench template & $\checkmark$ & $\checkmark$ & --- \\
C13 & \texttt{t1-\allowbreak{}datatable-\allowbreak{}drives-\allowbreak{}value} & CraftBench template & $\checkmark$ & $\checkmark$ & --- \\
C14 & \texttt{t1-\allowbreak{}default-\allowbreak{}cube-\allowbreak{}mesh-\allowbreak{}actor} & CraftBench template & $\checkmark$ & $\checkmark$ & --- \\
C15 & \texttt{t1-\allowbreak{}extraction-\allowbreak{}volume-\allowbreak{}per-\allowbreak{}actor-\allowbreak{}trigger} & CraftBench template & $\checkmark$ & $\checkmark$ & --- \\
C16 & \texttt{t1-\allowbreak{}gameplay-\allowbreak{}tag-\allowbreak{}gate} & CraftBench template & $\checkmark$ & $\checkmark$ & --- \\
C17 & \texttt{t1-\allowbreak{}movement-\allowbreak{}component-\allowbreak{}drives-\allowbreak{}actor} & CraftBench template & $\checkmark$ & $\checkmark$ & --- \\
C18 & \texttt{t1-\allowbreak{}mud-\allowbreak{}wade-\allowbreak{}cpp} & Third Person & $\checkmark$ & $\checkmark$ & --- \\
C19 & \texttt{t1-\allowbreak{}overlap-\allowbreak{}logs-\allowbreak{}once} & CraftBench template & $\checkmark$ & $\checkmark$ & --- \\
C20 & \texttt{t1-\allowbreak{}overlap-\allowbreak{}teleport-\allowbreak{}portal} & Third Person & $\checkmark$ & $\checkmark$ & --- \\
C21 & \texttt{t1-\allowbreak{}physics-\allowbreak{}drop-\allowbreak{}and-\allowbreak{}rest} & CraftBench template & $\checkmark$ & $\checkmark$ & --- \\
C22 & \texttt{t1-\allowbreak{}screen-\allowbreak{}tint-\allowbreak{}cpp} & Third Person & $\checkmark$ & $\checkmark$ & --- \\
C23 & \texttt{t2-\allowbreak{}gravity-\allowbreak{}floating-\allowbreak{}pawn-\allowbreak{}movement} & CraftBench template & $\checkmark$ & $\checkmark$ & --- \\
C24 & \texttt{t2-\allowbreak{}homing-\allowbreak{}projectile} & CraftBench template & $\checkmark$ & $\checkmark$ & --- \\
C25 & \texttt{t2-\allowbreak{}hud-\allowbreak{}layout-\allowbreak{}and-\allowbreak{}countdown} & Third Person & $\checkmark$ & $\checkmark$ & --- \\
C26 & \texttt{t2-\allowbreak{}ladder-\allowbreak{}climb-\allowbreak{}volume} & Third Person & $\checkmark$ & $\checkmark$ & --- \\
C27 & \texttt{t2-\allowbreak{}melee-\allowbreak{}ability-\allowbreak{}with-\allowbreak{}cooldown} & CraftBench template & $\checkmark$ & $\checkmark$ & --- \\
C28 & \texttt{t2-\allowbreak{}npc-\allowbreak{}follows-\allowbreak{}player} & Third Person & $\checkmark$ & $\checkmark$ & --- \\
C29 & \texttt{t2-\allowbreak{}race-\allowbreak{}clock-\allowbreak{}cpp} & Third Person & $\checkmark$ & $\checkmark$ & --- \\
C30 & \texttt{t2-\allowbreak{}timeline-\allowbreak{}color-\allowbreak{}cycle} & CraftBench template & $\checkmark$ & $\checkmark$ & --- \\
C31 & \texttt{t2-\allowbreak{}weapon-\allowbreak{}fire-\allowbreak{}animation-\allowbreak{}on-\allowbreak{}trigger} & Third Person & $\checkmark$ & $\checkmark$ & --- \\
C32 & \texttt{t3-\allowbreak{}gate-\allowbreak{}and-\allowbreak{}door-\allowbreak{}cpp} & Third Person & $\checkmark$ & $\checkmark$ & --- \\
C33 & \texttt{tp2-\allowbreak{}sprint-\allowbreak{}stamina} & Third Person & $\checkmark$ & $\checkmark$ & --- \\
\addlinespace[4pt]
\multicolumn{6}{@{}l}{\textbf{Blueprint/native-asset tasks}}\\[2pt]
B01 & \texttt{gp-\allowbreak{}additem-\allowbreak{}stack-\allowbreak{}fix-\allowbreak{}bp} & Third Person & $\checkmark$ & $\checkmark$ & --- \\
B02 & \texttt{gp-\allowbreak{}door-\allowbreak{}hitch-\allowbreak{}fix-\allowbreak{}bp} & Third Person & $\checkmark$ & $\checkmark$ & --- \\
B03 & \texttt{gp-\allowbreak{}dot-\allowbreak{}aoe-\allowbreak{}burn-\allowbreak{}bp} & Third Person & $\checkmark$ & $\checkmark$ & $\checkmark$ \\
B04 & \texttt{gp-\allowbreak{}double-\allowbreak{}jump-\allowbreak{}stamina-\allowbreak{}bp} & Third Person & $\checkmark$ & $\checkmark$ & $\checkmark$ \\
B05 & \texttt{gp-\allowbreak{}glide-\allowbreak{}stamina-\allowbreak{}bp} & Third Person & $\checkmark$ & $\checkmark$ & $\checkmark$ \\
B06 & \texttt{gp-\allowbreak{}heal-\allowbreak{}over-\allowbreak{}time-\allowbreak{}bp} & Third Person & $\checkmark$ & $\checkmark$ & $\checkmark$ \\
B07 & \texttt{gp-\allowbreak{}health-\allowbreak{}attribute-\allowbreak{}ops-\allowbreak{}bp} & Third Person & $\checkmark$ & $\checkmark$ & $\checkmark$ \\
B08 & \texttt{gp-\allowbreak{}poison-\allowbreak{}dot-\allowbreak{}stack-\allowbreak{}bp} & Third Person & $\checkmark$ & $\checkmark$ & $\checkmark$ \\
B09 & \texttt{t0-\allowbreak{}sanity-\allowbreak{}bp-\allowbreak{}log-\allowbreak{}on-\allowbreak{}beginplay} & CraftBench template & $\checkmark$ & $\checkmark$ & --- \\
B10 & \texttt{t1-\allowbreak{}blueprint-\allowbreak{}event-\allowbreak{}to-\allowbreak{}action} & CraftBench template & $\checkmark$ & $\checkmark$ & --- \\
B11 & \texttt{t1-\allowbreak{}blueprint-\allowbreak{}graph-\allowbreak{}on-\allowbreak{}beginplay} & CraftBench template & $\checkmark$ & $\checkmark$ & --- \\
B12 & \texttt{t1-\allowbreak{}dawn-\allowbreak{}fog-\allowbreak{}lighting-\allowbreak{}rig} & Third Person & $\checkmark$ & --- & $\checkmark$ \\
B13 & \texttt{t1-\allowbreak{}hero-\allowbreak{}blueprint-\allowbreak{}copy-\allowbreak{}with-\allowbreak{}flashlight} & Third Person & $\checkmark$ & --- & $\checkmark$ \\
B14 & \texttt{t1-\allowbreak{}mud-\allowbreak{}wade-\allowbreak{}bp} & Third Person & $\checkmark$ & $\checkmark$ & $\checkmark$ \\
B15 & \texttt{t1-\allowbreak{}playable-\allowbreak{}level-\allowbreak{}bootstrap} & Third Person & $\checkmark$ & --- & $\checkmark$ \\
B16 & \texttt{t1-\allowbreak{}screen-\allowbreak{}tint-\allowbreak{}bp} & Third Person & $\checkmark$ & $\checkmark$ & $\checkmark$ \\
B17 & \texttt{t1-\allowbreak{}third-\allowbreak{}person-\allowbreak{}chase-\allowbreak{}camera} & Third Person & $\checkmark$ & --- & $\checkmark$ \\
B18 & \texttt{t1-\allowbreak{}walk-\allowbreak{}animation-\allowbreak{}footstep-\allowbreak{}cues} & Third Person & $\checkmark$ & --- & $\checkmark$ \\
B19 & \texttt{t2-\allowbreak{}consistent-\allowbreak{}enum-\allowbreak{}names} & Third Person & $\checkmark$ & --- & $\checkmark$ \\
B20 & \texttt{t2-\allowbreak{}cutscene-\allowbreak{}camera-\allowbreak{}push-\allowbreak{}and-\allowbreak{}hero-\allowbreak{}rise} & Third Person & $\checkmark$ & --- & $\checkmark$ \\
B21 & \texttt{t2-\allowbreak{}datatable-\allowbreak{}csv-\allowbreak{}export} & Third Person & $\checkmark$ & --- & $\checkmark$ \\
B22 & \texttt{t2-\allowbreak{}race-\allowbreak{}clock-\allowbreak{}bp} & Third Person & $\checkmark$ & $\checkmark$ & $\checkmark$ \\
B23 & \texttt{t2-\allowbreak{}weapon-\allowbreak{}held-\allowbreak{}in-\allowbreak{}right-\allowbreak{}hand} & Third Person & $\checkmark$ & --- & $\checkmark$ \\
B24 & \texttt{t3-\allowbreak{}gate-\allowbreak{}and-\allowbreak{}door-\allowbreak{}bp} & Third Person & $\checkmark$ & $\checkmark$ & $\checkmark$ \\
B25 & \texttt{t3-\allowbreak{}piercing-\allowbreak{}projectile} & Third Person & $\checkmark$ & --- & $\checkmark$ \\
\addlinespace[4pt]
\multicolumn{6}{@{}l}{\textbf{Editor-scripting tasks}}\\[2pt]
P01 & \texttt{kp-\allowbreak{}anim-\allowbreak{}track-\allowbreak{}bake} & Third Person & $\checkmark$ & --- & $\checkmark$ \\
P02 & \texttt{kp-\allowbreak{}blueprint-\allowbreak{}actor-\allowbreak{}audit-\allowbreak{}report} & Third Person & $\checkmark$ & --- & $\checkmark$ \\
P03 & \texttt{kp-\allowbreak{}character-\allowbreak{}boom-\allowbreak{}and-\allowbreak{}movement} & Third Person & $\checkmark$ & --- & $\checkmark$ \\
P04 & \texttt{kp-\allowbreak{}config-\allowbreak{}source-\allowbreak{}audit} & Third Person & $\checkmark$ & --- & $\checkmark$ \\
P05 & \texttt{kp-\allowbreak{}derived-\allowbreak{}class-\allowbreak{}search} & Third Person & $\checkmark$ & --- & $\checkmark$ \\
P06 & \texttt{kp-\allowbreak{}engine-\allowbreak{}source-\allowbreak{}search} & Third Person & $\checkmark$ & --- & $\checkmark$ \\
P07 & \texttt{kp-\allowbreak{}fog-\allowbreak{}and-\allowbreak{}postprocess-\allowbreak{}rig} & Third Person & $\checkmark$ & --- & $\checkmark$ \\
P08 & \texttt{kp-\allowbreak{}motion-\allowbreak{}set-\allowbreak{}shares-\allowbreak{}one-\allowbreak{}rig} & Third Person & $\checkmark$ & --- & $\checkmark$ \\
P09 & \texttt{kp-\allowbreak{}retarget-\allowbreak{}maps-\allowbreak{}two-\allowbreak{}rigs} & Third Person & $\checkmark$ & --- & $\checkmark$ \\
P10 & \texttt{kp-\allowbreak{}routine-\allowbreak{}usage-\allowbreak{}search} & Third Person & $\checkmark$ & --- & $\checkmark$ \\
P11 & \texttt{kp-\allowbreak{}spawn-\allowbreak{}level-\allowbreak{}actors} & Third Person & $\checkmark$ & --- & $\checkmark$ \\
P12 & \texttt{kp-\allowbreak{}typed-\allowbreak{}input-\allowbreak{}bindings} & Third Person & $\checkmark$ & --- & $\checkmark$ \\
\addlinespace[4pt]
\end{longtable}
\endgroup

\begin{table}[!ht]
\centering\small
\setlength{\tabcolsep}{4pt}
\begin{tabularx}{\linewidth}{@{}L{.20\linewidth}XL{.12\linewidth}@{}}
\toprule
Task family & Requested behavior & Catalog rows \\
\midrule
Area burn & Apply periodic damage inside a bounded area and stop after its duration. & C02 / B03 \\
Double jump & Reverse a fall with an upward impulse; debit Power once and reject insufficient Power. & C03 / B04 \\
Glide & Slow descent while draining Power and restore normal falling when Power is exhausted. & C04 / B05 \\
Healing over time & Restore Health repeatedly for a limited duration without exceeding MaxHealth. & C06 / B06 \\
Health operations & Initialize writable Health and provide repeatable damage and healing abilities. & C07 / B07 \\
Stacking poison & Apply periodic damage with up to three stacks and refresh the duration on reapplication. & C09 / B08 \\
Mud wading & Slow movement and change animation in mud, restore both afterward, and preserve the unaffected character. & C18 / B14 \\
Speed-driven tint & Track movement speed relative to the current maximum while preserving an unrelated screen setting. & C22 / B16 \\
Race clock & Run a timed coin-collection round with live readouts, final scoring, and restart behavior. & C29 / B22 \\
Gates and door & Open each new gate for its required pair of named crates while preserving the existing door behavior. & C32 / B24 \\
\bottomrule
\end{tabularx}
\caption{The ten gameplay specifications with C++ and Blueprint counterparts. Each pair names the same starting project, runtime map, and test class in its task definitions. Both counterparts require build and runtime checks; the Blueprint counterpart also requires artifact inspection. These summaries describe the requested behavior; the task prompts specify the acceptance tolerances and deliverables.}
\label{tab:matched-task-design}
\end{table}

\clearpage
\section{Comparison with Game-Development Benchmarks}
\label{app:task-examples}
\label{app:benchmark-comparison}

Table~\ref{tab:design-comparison} compares what agents are asked to produce and the evidence used to evaluate it. Different task scopes support different claims: atomic gameplay checks, visual cutscene quality, and complete game generation are not interchangeable outcomes.

\begin{table}[!ht]
\centering\footnotesize
\setlength{\tabcolsep}{4pt}
\begin{tabularx}{\linewidth}{@{}L{.19\linewidth}L{.22\linewidth}YY@{}}
\toprule
Benchmark & Setting and task source & Agent's deliverable & Grading evidence \\
\midrule
GameDevBench\newline\citep{gamedevbench} & Godot; 333 tasks from web/video tutorials & Scripts, scenes, resources & In-engine tests of behavior and scene state \\
\addlinespace
GameEngineBench\newline\citep{gameenginebench} & Unreal; 110 tasks from nine projects & C++ changes in existing projects & Build and runtime tests, followed by an LLM behavioral judge \\
\addlinespace
OpenGameEval\newline\citep{opengameeval} & Roblox; expert-authored tasks (47 in the initial release) & Scripts and changes to existing experiences & Executable tests in Roblox Studio \\
\addlinespace
GBQA\newline\citep{gbqa} & Web games; 124 human-verified injected bugs in 30 agent-built games & Bug reports from interactive exploration & LLM critic matches reports to ground truth; recall of verified bugs \\
\addlinespace
CutsceneBench\newline\citep{cutsceneagent} & Unreal; 65 authored cutscene scenarios & Level Sequence assets & Tool and sequence checks; model-judged quality on 25 sampled videos per model \\
\addlinespace
AutoUE\newline\citep{autoue} & Unreal; complete game generation & Game projects with scenes and gameplay & Automated play-testing; LLM-judged game quality \\
\addlinespace
GameXpert-Bench\newline\citep{gamexpertbench} & No prescribed engine, browser JavaScript; 97 generation tasks, 100 repair tasks from 50 human-verified levels, 102 optimization turns & All game source files, written from a blank workspace; source patches for repair & Deterministic headless-browser probes (repair); evaluation-agent rubric with human rating (generation); hidden-rubric judging (optimization) \\
\addlinespace
\textbf{\benchname{}} & Unreal; \numtasks{} bounded gameplay/editor tasks & Source changes, native assets, editor-script outputs & Task-specific build, runtime, and saved-asset checks; no LLM judge \\
\bottomrule
\end{tabularx}
\caption{Benchmark designs, using the cited releases. The OpenGameEval count refers to its initial release; its task bank has since expanded. \benchname{} also includes \nummatchedpairs{} C++/BP gameplay pairs and evaluates two editor MCP configurations.}
\label{tab:design-comparison}
\end{table}

\end{document}